\documentclass{article}
\usepackage{spconf,amsmath,graphicx}

\usepackage{tabularx}
\usepackage{caption}
\usepackage{subcaption}
\graphicspath{{figs/}{./}} 
\usepackage{amsmath}

\usepackage{times}
\usepackage{epsfig}
\usepackage{amssymb}
\usepackage{booktabs}

\usepackage[nodisplayskipstretch]{setspace}
\usepackage{caption}
\title{Dynamic Multi-View Scene Reconstruction Using Neural Implicit Surface}
\name{Decai Chen$^1$,  Haofei Lu$^{1,2}$, Ingo Feldmann$^1$, Oliver Schreer$^1$, Peter Eisert$^{1,3}$}
\address{%
    $^1$Fraunhofer HHI \,
    $^2$TU Berlin \,
    $^3$HU Berlin \,
}

\begin{document}
\ninept

\maketitle
\begin{abstract}
Reconstructing general dynamic scenes is important for many computer vision and graphics applications. Recent works represent the dynamic scene with neural radiance fields for photorealistic view synthesis, while their surface geometry is under-constrained and noisy. Other works introduce surface constraints to the implicit neural representation to disentangle the ambiguity of geometry and appearance field for static scene reconstruction. To bridge the gap between rendering dynamic scenes and recovering static surface geometry, we propose a template-free method to reconstruct surface geometry and appearance using neural implicit representations from multi-view videos. We leverage topology-aware deformation and the signed distance field to learn complex dynamic surfaces via differentiable volume rendering without scene-specific prior knowledge like template models. Furthermore, we propose a novel mask-based ray selection strategy to significantly boost the optimization on challenging time-varying regions. Experiments on different multi-view video datasets demonstrate that our method achieves high-fidelity surface reconstruction as well as photorealistic novel view synthesis.
\end{abstract}
%

\section{Introduction}
Recent success of deep learning techniques in 2D domain has sparked a surge of interest in higher dimensional problems, such as 3D computer vision tasks and their application fields in 
VR/AR, 
movie production, games etc. 
For instance, the neural radiance field (NeRF)~\cite{nerf} demonstrates that Multi-Layer Perceptron (MLP) neural networks can represent a scene by implicitly encoding the geometry and appearance into the parameter of networks. The learned model allows free-viewpoint photorealistic novel view synthesis through traditional volume rendering techniques. 

While NeRF was initially designed for static content, subsequent dynamic works 
either condition the neural field with addtional temporal input~\cite{nsff,xian2021space,nrf4}, or jointly optimize a deformation field and a neural radiance network~\cite {d-nerf, nerfies, NR-NeRF}. Nevertheless, compared to remarkable achievements in view synthesis, their performance
in geometry representation is relatively unsatisfying.
To address this problem, several works~\cite{volsdf,neus,NeuralWarp,dneus} employ the signed distance function(SDF) for implicit surface representation. Compared to density, SDF can be more efficiently regularized, relieving the entanglement between shape and appearance.
However, these methods only reconstruct static scenes instead of more commonly seen dynamic ones.

Another closely related research topic is recovering dynamic humans from videos using an articulated human body model such as SMPL~\cite{SMPL:2015}.
Although these methods~\cite{anerf, peng2022animatable, chen2021accurate, neuralactor,peng2021neural,dd-nerf} have demonstrated impressive efficacy in view synthesis and reconstruction in terms of the human body, the requirement of human priors like the SMPL model hinders them from generalizing to other dynamic scenes. General dynamic scenes are challenging but widely used in applications, such as interaction with props and objects like balls, loose garments (skirts, cloaks), and movements of animals or robots.

To overcome the above limitations, we propose DySurf, a neural multi-view dynamic surface reconstruction method without prior knowledge of the target shape. We map the spatial points along a camera ray from observation space to the canonical space
, which is explicitly parameterized by 
the SE(3) field computed from a deformation network. This enables the model to share the coherence of geometry and appearance information across time. To handle challenging topology changes in dynamic scenes, we adopt the design of a hyperdimensional network~\cite {hypernerf}.
For rendering, we employ the neural SDF and radiance field to represent geometry and appearance in the canonical space. To recover dynamic areas with fast motion, we propose a novel ray selection strategy that assigns a higher sampling probability to pixels of interest derived from 
dynamic masks.

We show the performance of our method across different scenarios on a public dataset (GeneBody~\cite {cheng2022genebody}) as well as a multi-view dataset captured by ourselves. With the focus on dynamic surface reconstruction, we also demonstrate the capability of our work on view synthesis.
In summary, the main contributions of this work are as follows: 1) An end-to-end framework called DySurf for topology-aware dynamic surface reconstruction from multi-view videos without templates; 2) A novel mask-based ray selection strategy to boost the optimization by focusing more on the time-varying foreground region; 3) Extensive evaluation of high-fidelity surface reconstruction on the public GeneBody dataset as well as our captured dataset.

\begin{figure}
\begin{center}
\includegraphics[trim={4.8cm 9.8cm 4.1cm 2.8cm},clip,width=\linewidth]{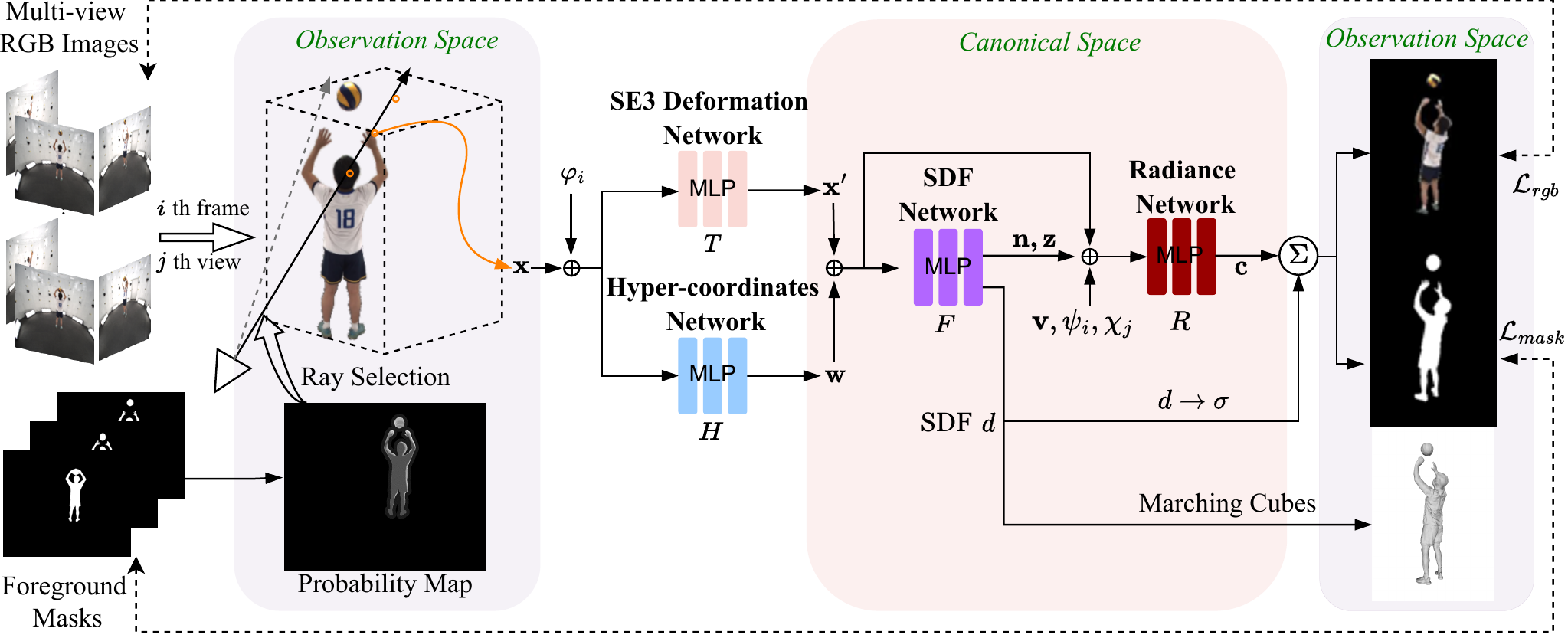}
\end{center}
\caption{Overview of our approach.}
\label{overview}
\end{figure}

\section{Method}
The overview of our method is demonstrated in Figure~\ref{overview}. 
Our goal is to reconstruct high-fidelity time-varying surfaces from multi-view videos. The inputs are multi-view RGB image sequences $\{\mathit{I_{i,j}}, \mathit{S_{i,j}} : i\in [1,N], j\in [1,M]\}$ containing $N$ frames and $M$ views, where $\mathit{I_{i,j}}$ is the color image of the $i$-th frame from the $j$-th view and $\mathit{S_{i,j}}$ is the corresponding object segmentation mask. Masks can be obtained by off-the-shelf foreground-background segmentation framework, e.g., BackgroundMattingV2~\cite{BGMv2}. Camera parameters including intrinsics and extrinsics are also provided. They are typically calculated from a dedicated calibration process. Given the above inputs, we aim to create a time-coherent 4D representations of the dynamic scene, where we can recover high-quality geometry surfaces and synthesize realistic novel view images.


\subsection{Neural Dynamic Surface Representation} \label{representation}

\noindent
\textbf{Deformation Field}. In deformation-based dynamic scene representation, it is important to properly define the connection between the time-varying observation space where the cameras capture the scene, and the canonical space where the underlying geometry and appearance can be queried. 
Given the learnable per-frame latent deformation code $\mathbf{\varphi}_i$, we model the spatial deformation using a SE(3) field network~\cite{nerfies} $T: (\mathbf{x},\mathbf{\varphi}_i) \rightarrow (\hat{\mathbf{r}},\hat{\mathbf{t}})$, where $\hat{\mathbf{r}},\hat{\mathbf{t}}\in \mathbb{R}^6$ is a 6-DOF vector in the continuous SE(3) field parameterizing the rotation and translation, respectively. Inspired by HyperNeRF~\cite{hypernerf}, we utilize a hyper-coordinates network $H: \mathbf{(x,\varphi}_{i}) \rightarrow \mathbf{w} \in \mathbb{R}^{m}$ to gain more flexibility for representing various topologies of dynamic scene surfaces, by extending the conventional 3D canonical space with additional $m$ dimensions. To summarize, the mapping from a 3D sampled point in the observation space (also known as deformed space) $\mathbf{x}$ to the hyper canonical-space coordinates $(\mathbf{x',w})$ is defined as: 
\begin{equation}
    (T\mathbf{(x,\varphi}_i), H\mathbf{(x,\varphi}_i)) \rightarrow (\mathbf{x',w}) \in \mathbb{R}^{3+m}.
\end{equation}

\noindent
\textbf{Neural SDF and Radiance Field}. In this work, we model the surface geometry using an SDF network 
$F: (\mathbf{x',w}) \rightarrow (d, \mathbf{z})$,
where $\mathbf{z} \in \mathbb{R}^{q} $ is the geometry feature to condition the radiance network $R$. 
In addition, 
we calculate the surface normal using gradient from auto-differentiation $\mathbf{n} = \nabla F(\mathbf{x', w})$, to better disentangle the geometry and appearance fields.

For volume rendering, we follow VolSDF~\cite{volsdf} to transform the SDF value $d$ to the density $\sigma$ used in volume integration.

To offset the variation of illumination and exposure across different input frames and camera views, we additionally condition appearance codes $\psi_i$ and $\chi_j$ to the $i$-th frame and the $j$-th view. Finally, the radiance network $R$ can be formulated as:
\begin{equation}
    R(\mathbf{x'},\mathbf{w},\mathbf{n},\mathbf{z},\mathbf{v},\psi_i,\chi_j) \rightarrow \mathbf{c}.
\label{color}
\end{equation}
To approximate the ray rendering integral via discretization, we sample $N_s$ points along each ray based on an error bound for the opacity approximation~\cite{volsdf}.
Finally, the volume rendered color for a pixel is obtained by alpha blending over all sampled colors $\mathbf{c}$ along the ray.


\begin{figure}[t]
     \centering
    \hspace*{\fill}
     \begin{subfigure}[b]{0.48\columnwidth}
         \centering
         \includegraphics[trim={0 0cm 0 0cm},clip, width=\textwidth]{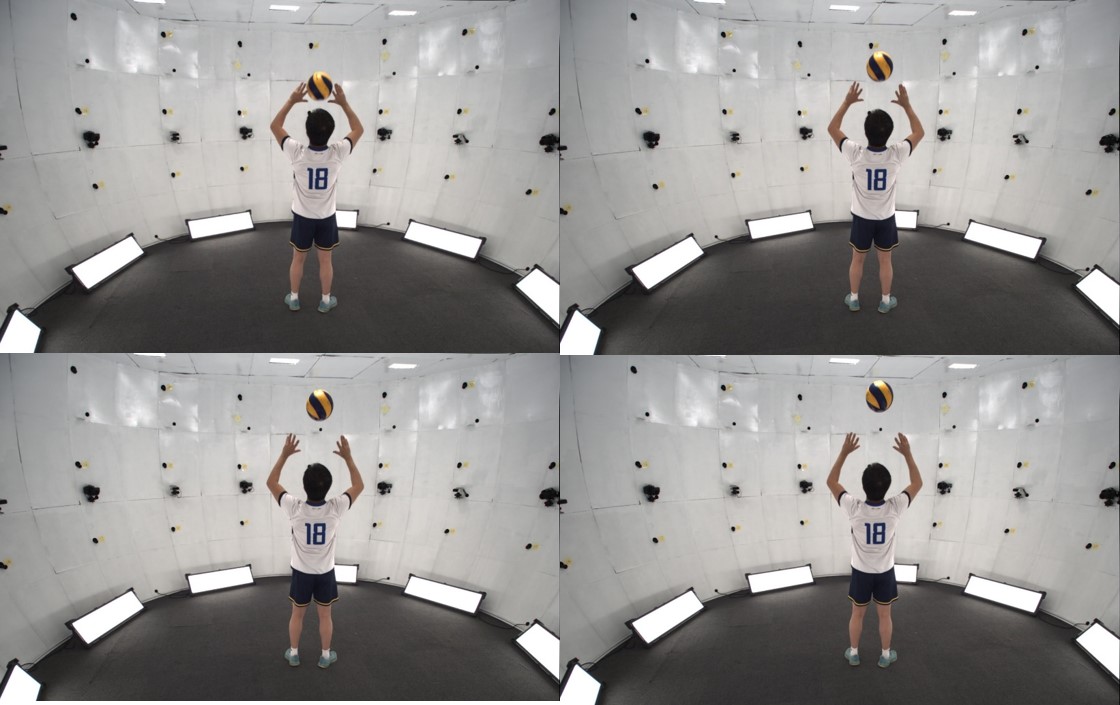}
         \caption{Input image examples.}
         \label{fig:img}
     \end{subfigure}
    \hspace*{\fill}
     \begin{subfigure}[b]{0.48\columnwidth}
         \centering
         \includegraphics[trim={0 2cm 0 0},clip, width=\textwidth]{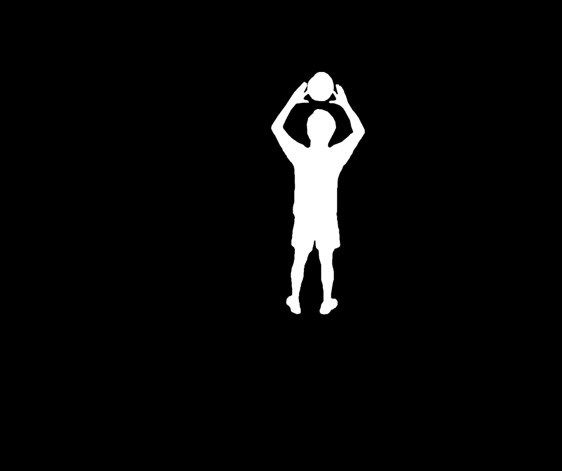}
         \caption{Mask at the specified frame.}
         \label{fig:mask}
     \end{subfigure}
    \hspace*{\fill}
    
    \hspace*{\fill}
     \begin{subfigure}[b]{0.48\columnwidth}
         \centering
         \includegraphics[trim={0 2cm 0 0},clip, width=\textwidth]{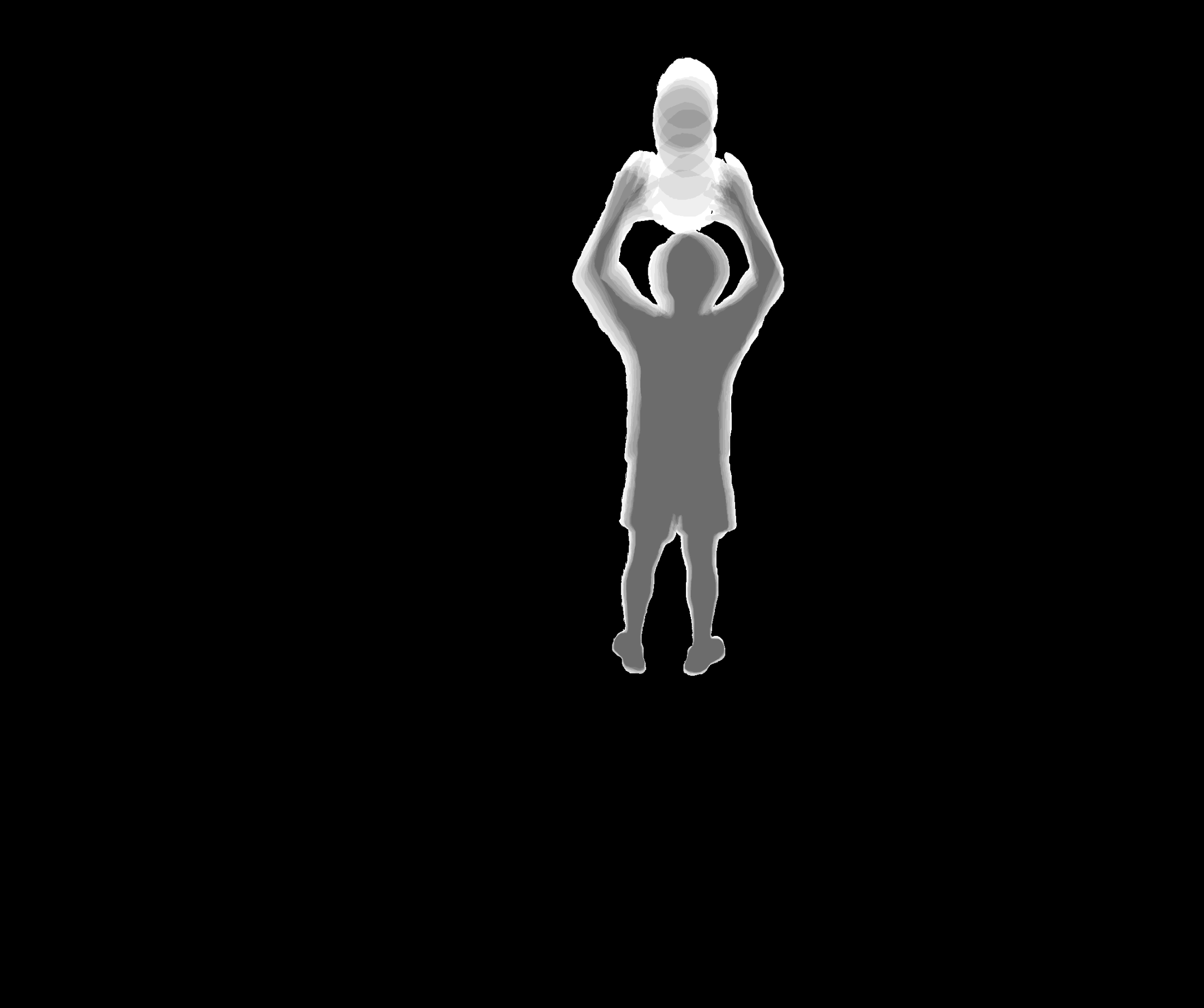}
         \caption{Dynamic motion map over frames.}
         \label{fig:motion}
     \end{subfigure}
    \hspace*{\fill}
     \begin{subfigure}[b]{0.48\columnwidth}
         \centering
         \includegraphics[trim={0 2cm 0 0},clip, width=\textwidth]{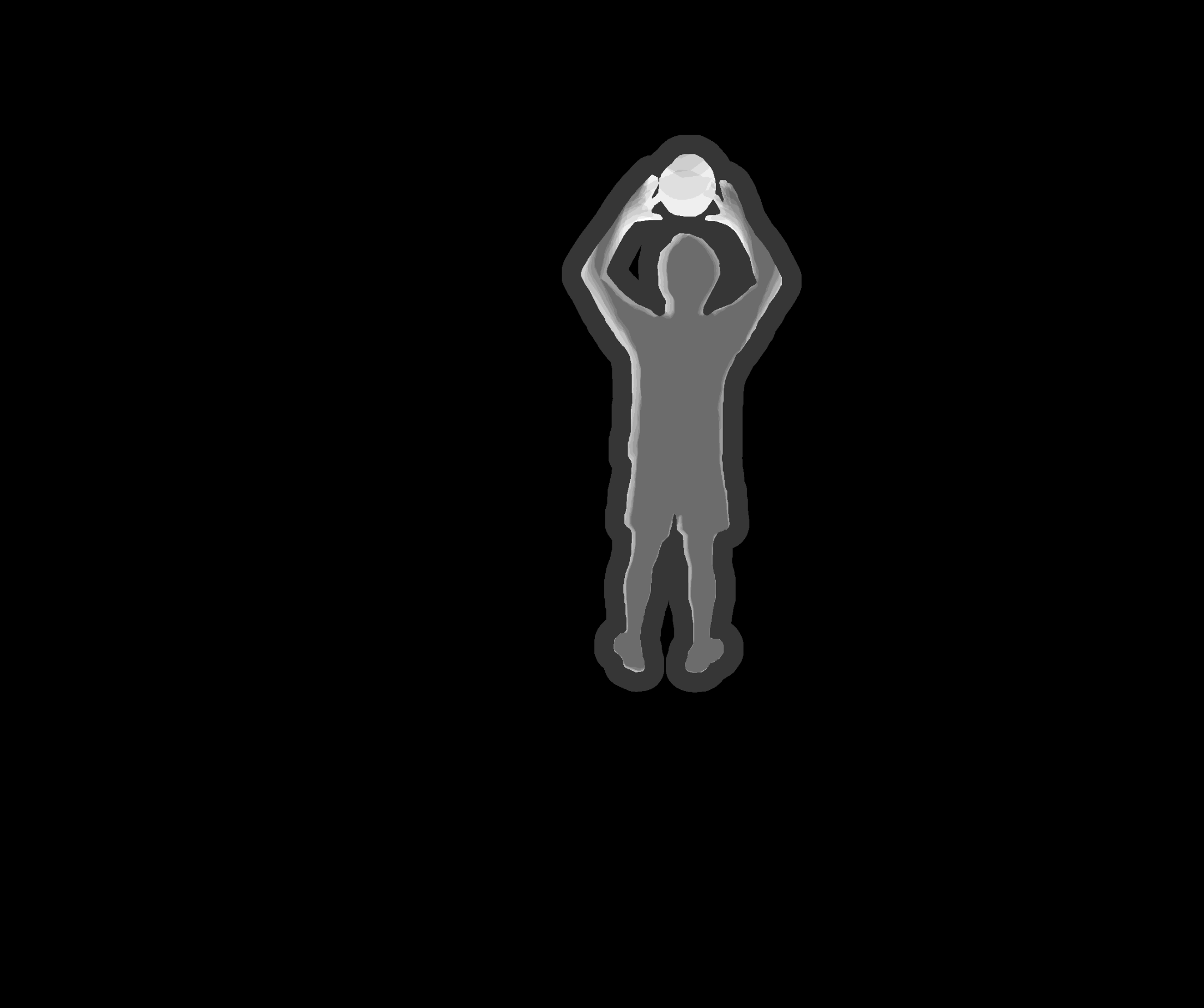}
         \caption{Final probability map at the \\ specified frame.}
         \label{fig:prob_map}
     \end{subfigure}
    \hspace*{\fill}
\caption{Development of probability map for ray selection.}
\label{fig:ray_selec}
\end{figure}

\begin{figure*}[t]
    \centering
    \includegraphics[width=0.162\linewidth]{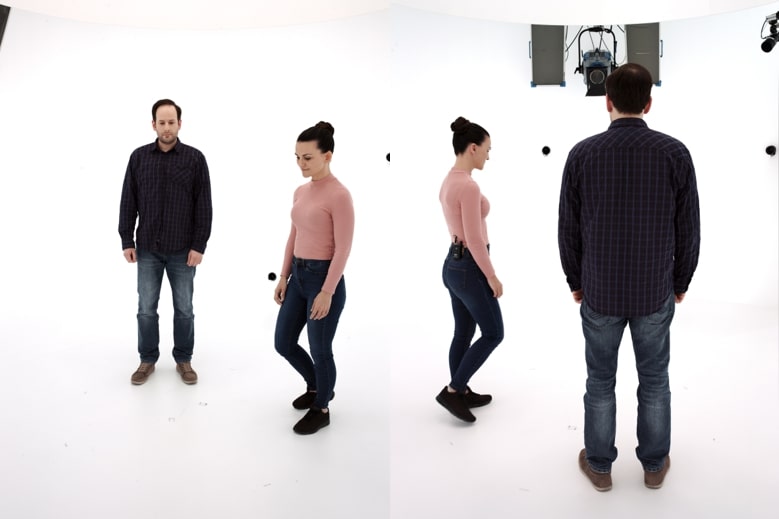}
    \includegraphics[width=0.162\linewidth]{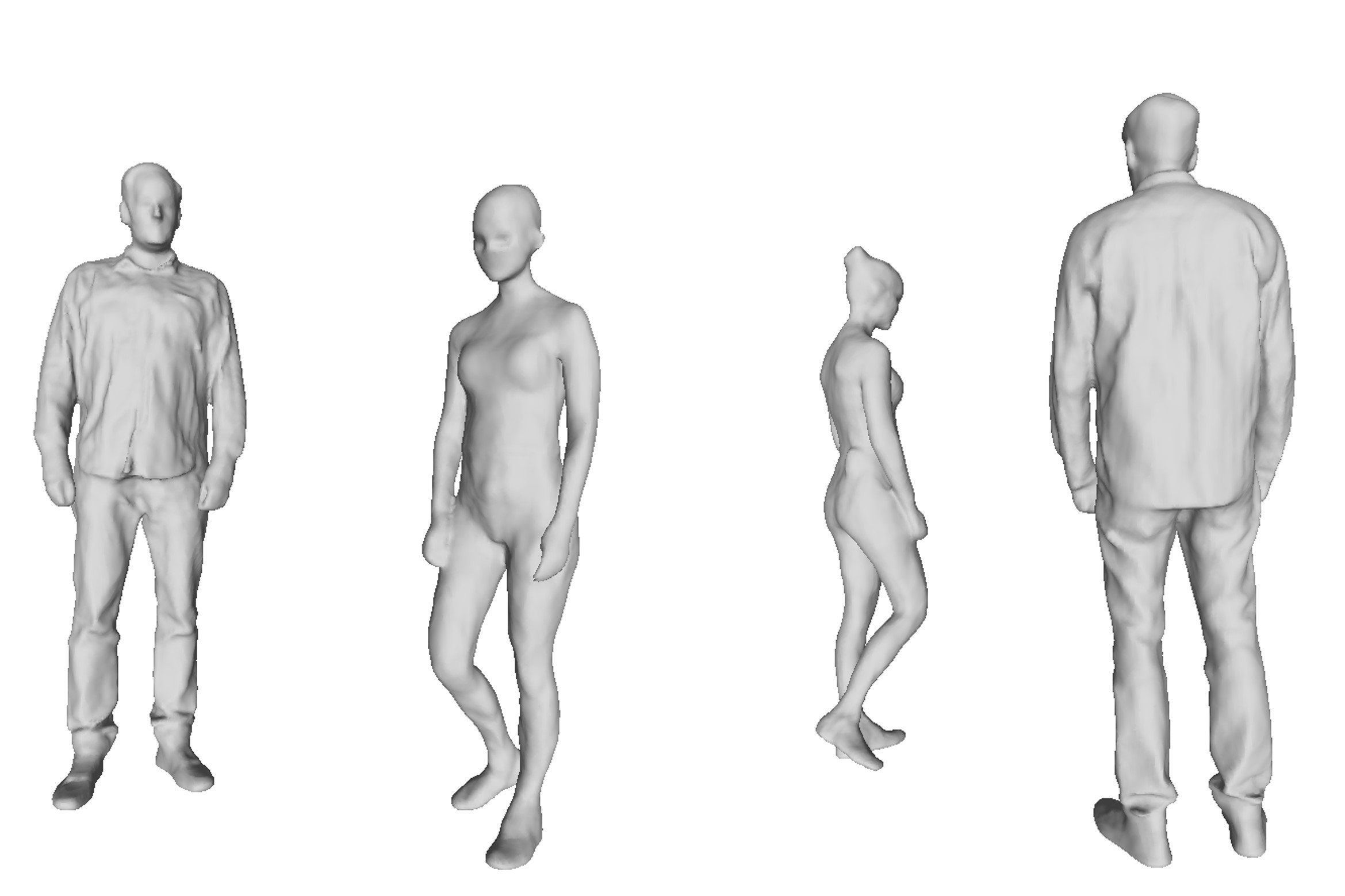}
    \includegraphics[width=0.162\linewidth]{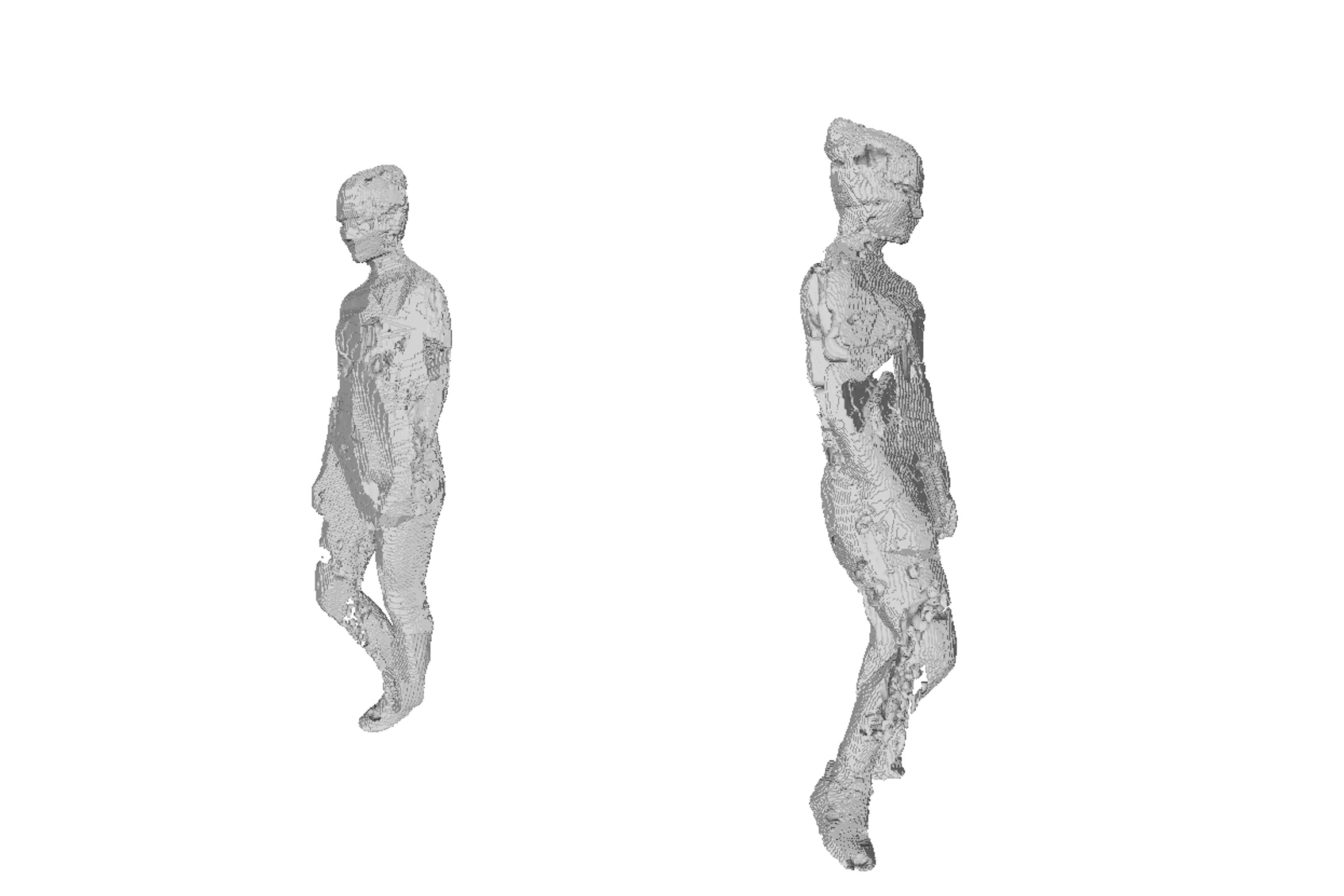}
    \includegraphics[width=0.162\linewidth]{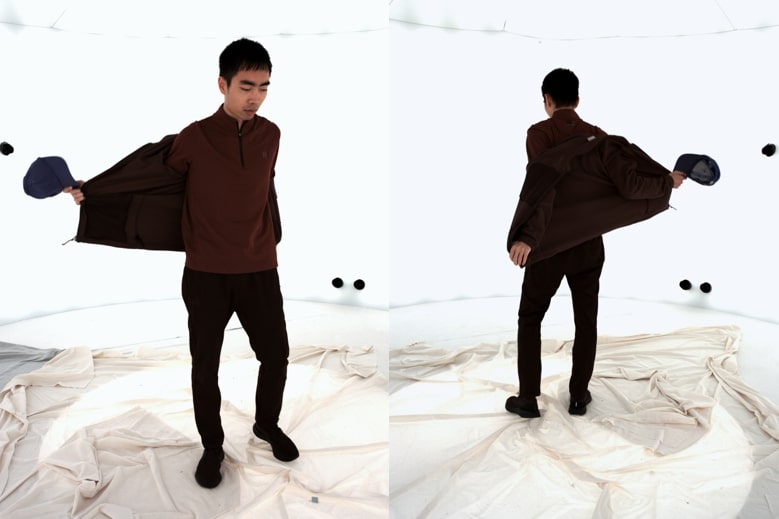}
    \includegraphics[width=0.162\linewidth]{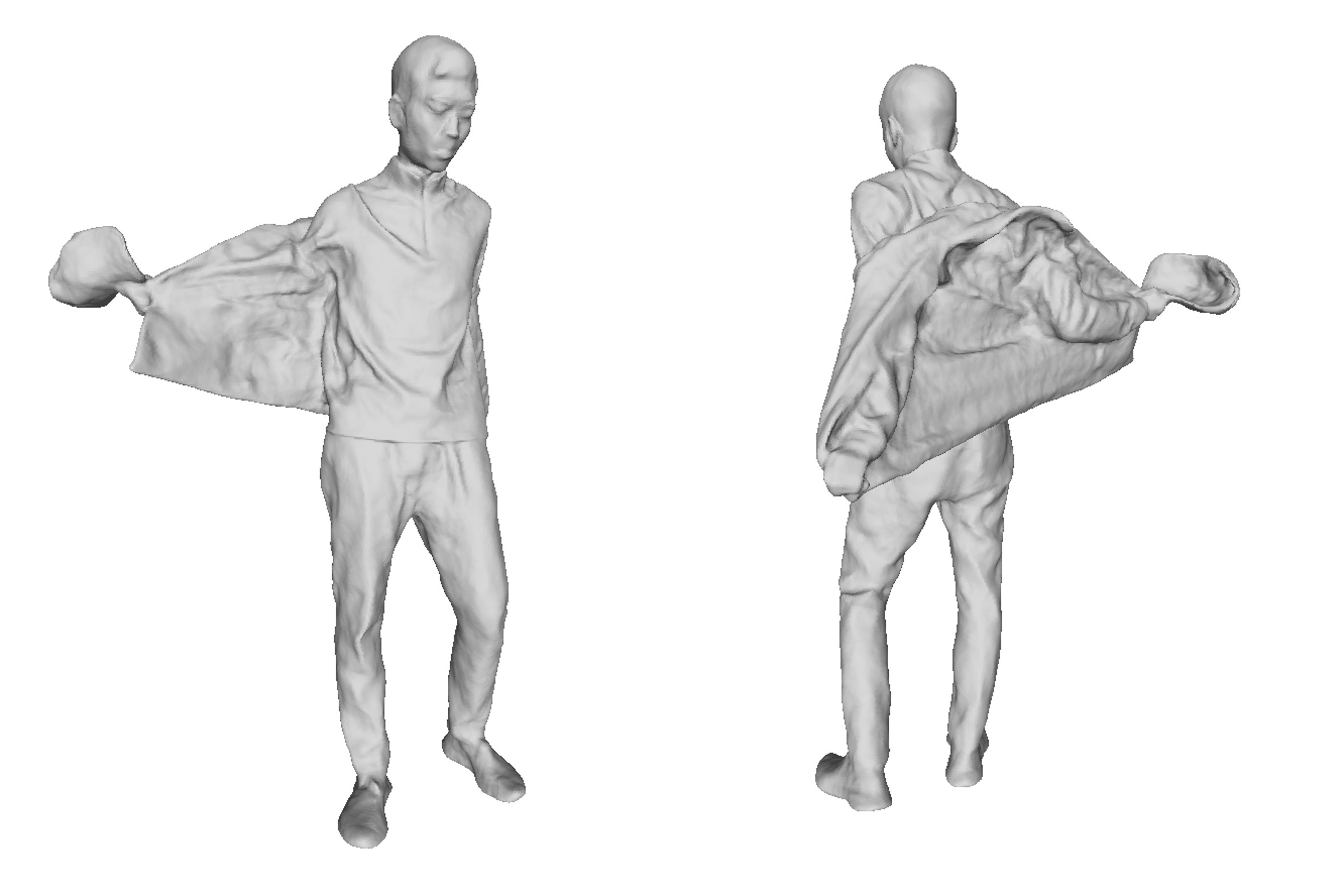}
    \includegraphics[width=0.162\linewidth]{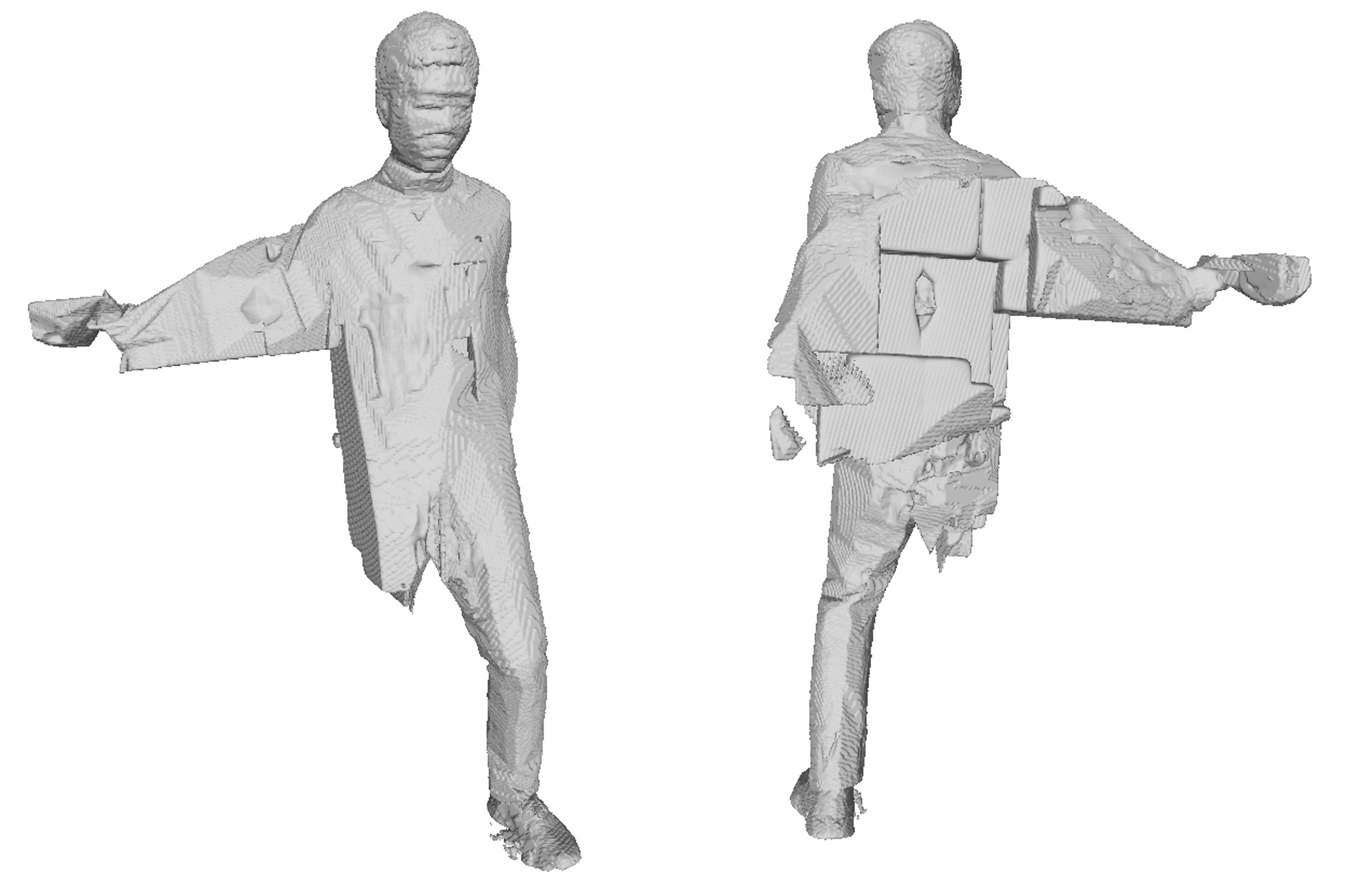}
    
    \includegraphics[width=0.162\linewidth]{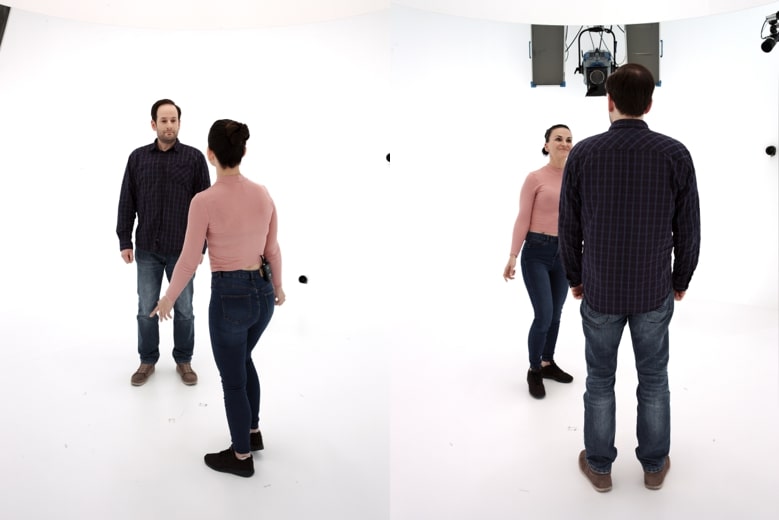}
    \includegraphics[width=0.162\linewidth]{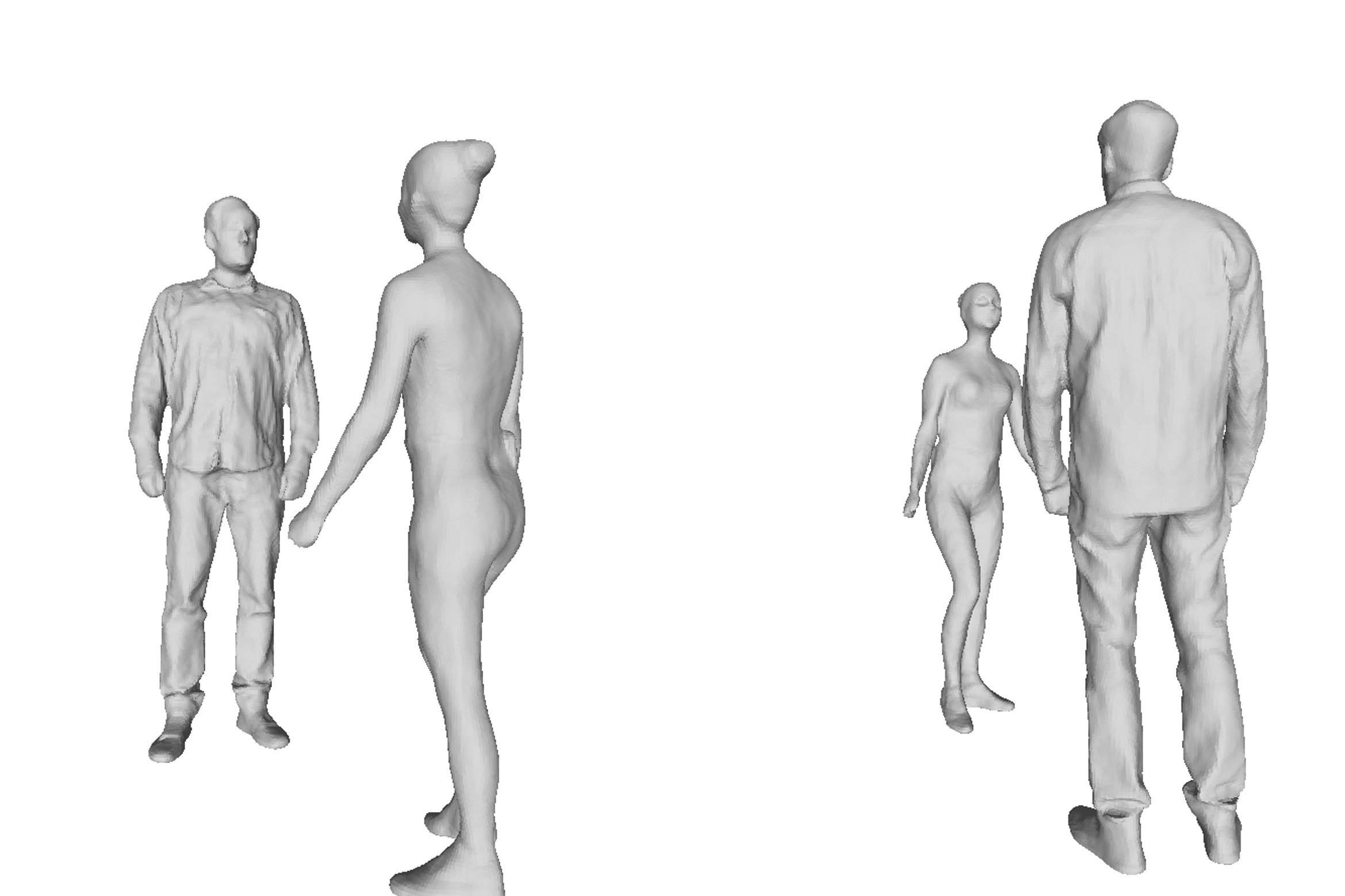}
    \includegraphics[width=0.162\linewidth]{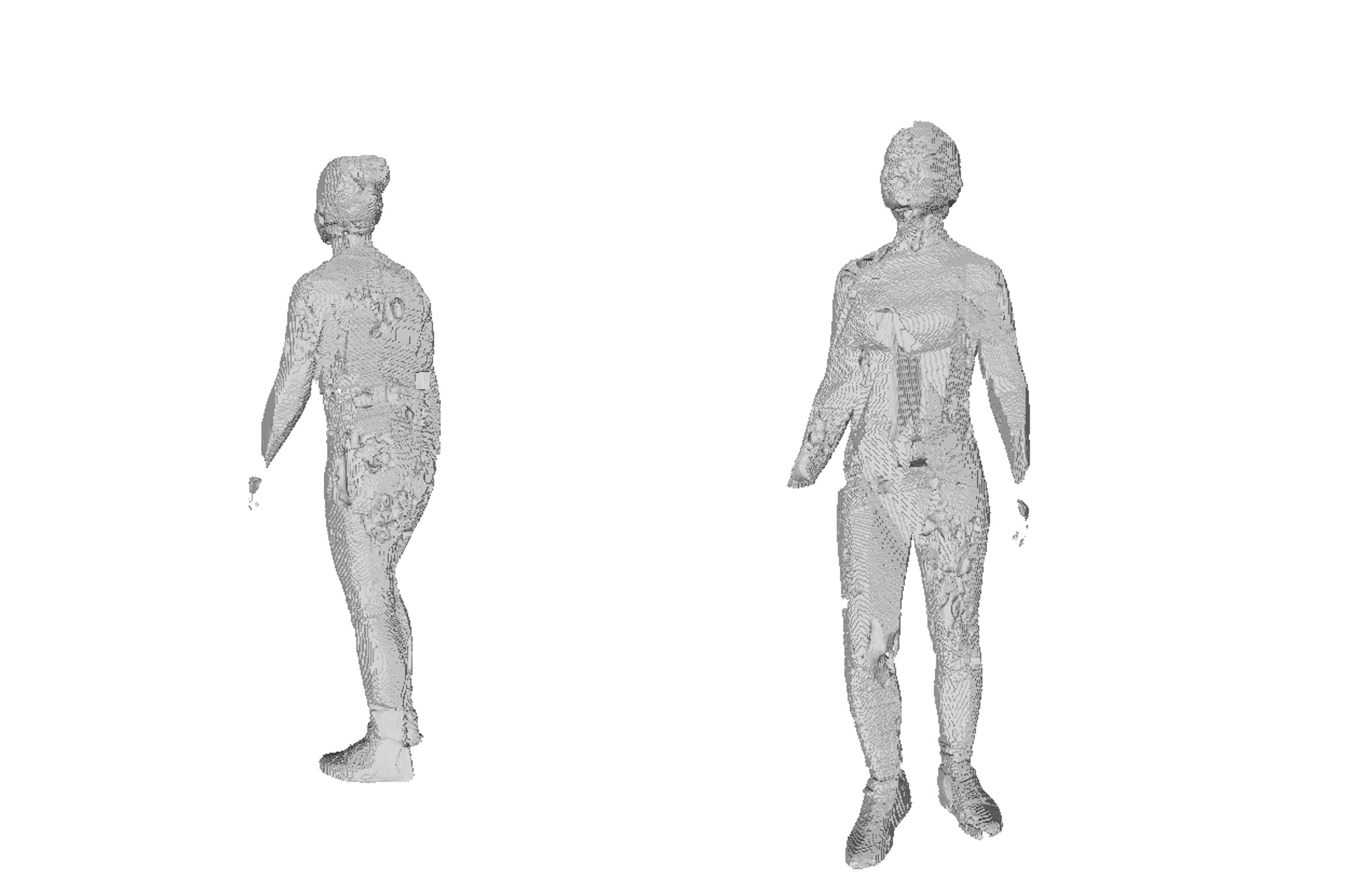}
    \includegraphics[width=0.162\linewidth]{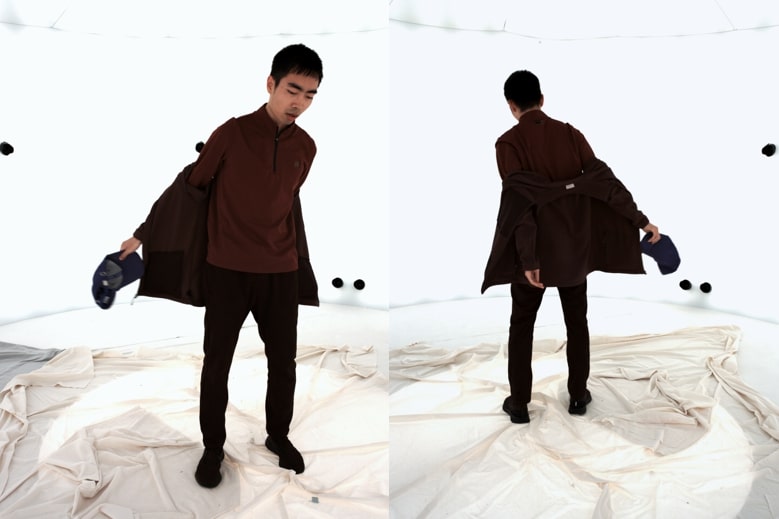}
    \includegraphics[width=0.162\linewidth]{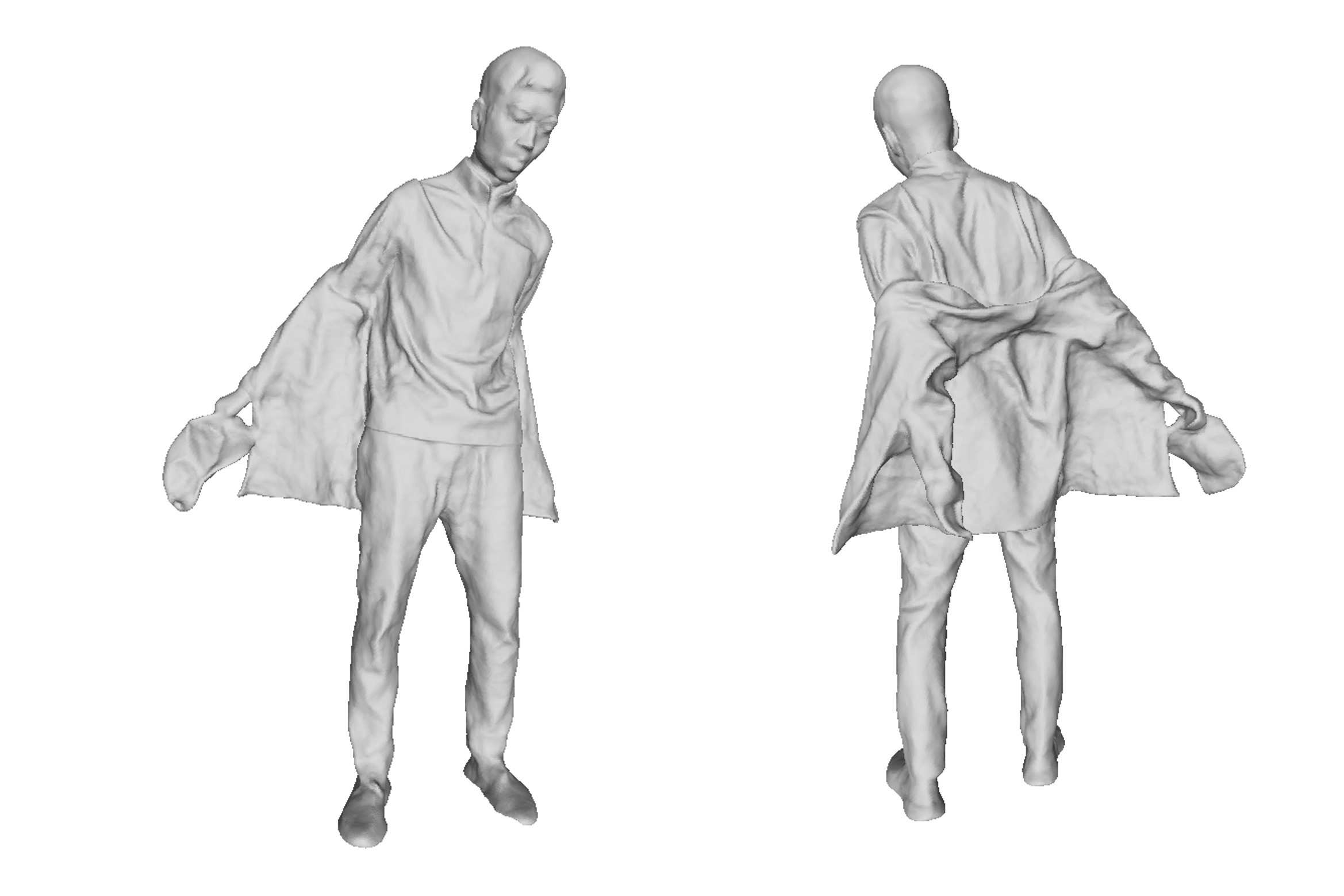}
    \includegraphics[width=0.162\linewidth]{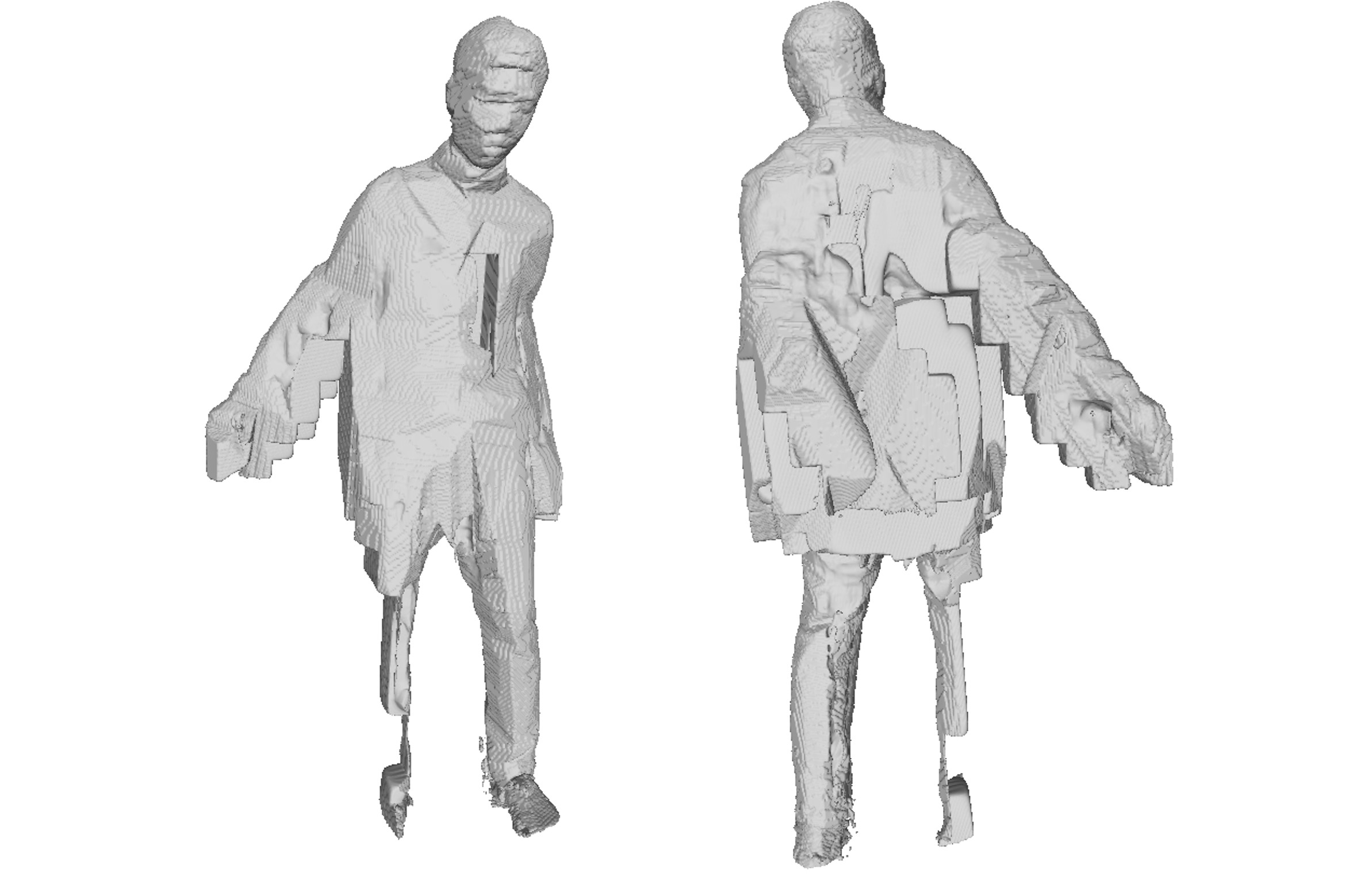}
    
    
    \includegraphics[width=0.162\linewidth]{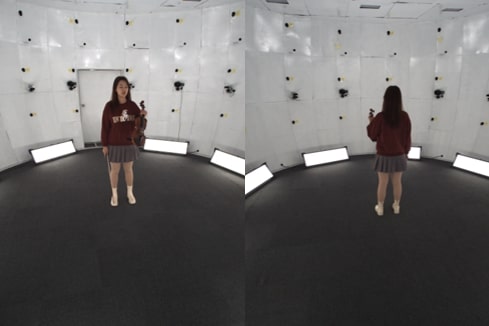}
    \includegraphics[width=0.162\linewidth]{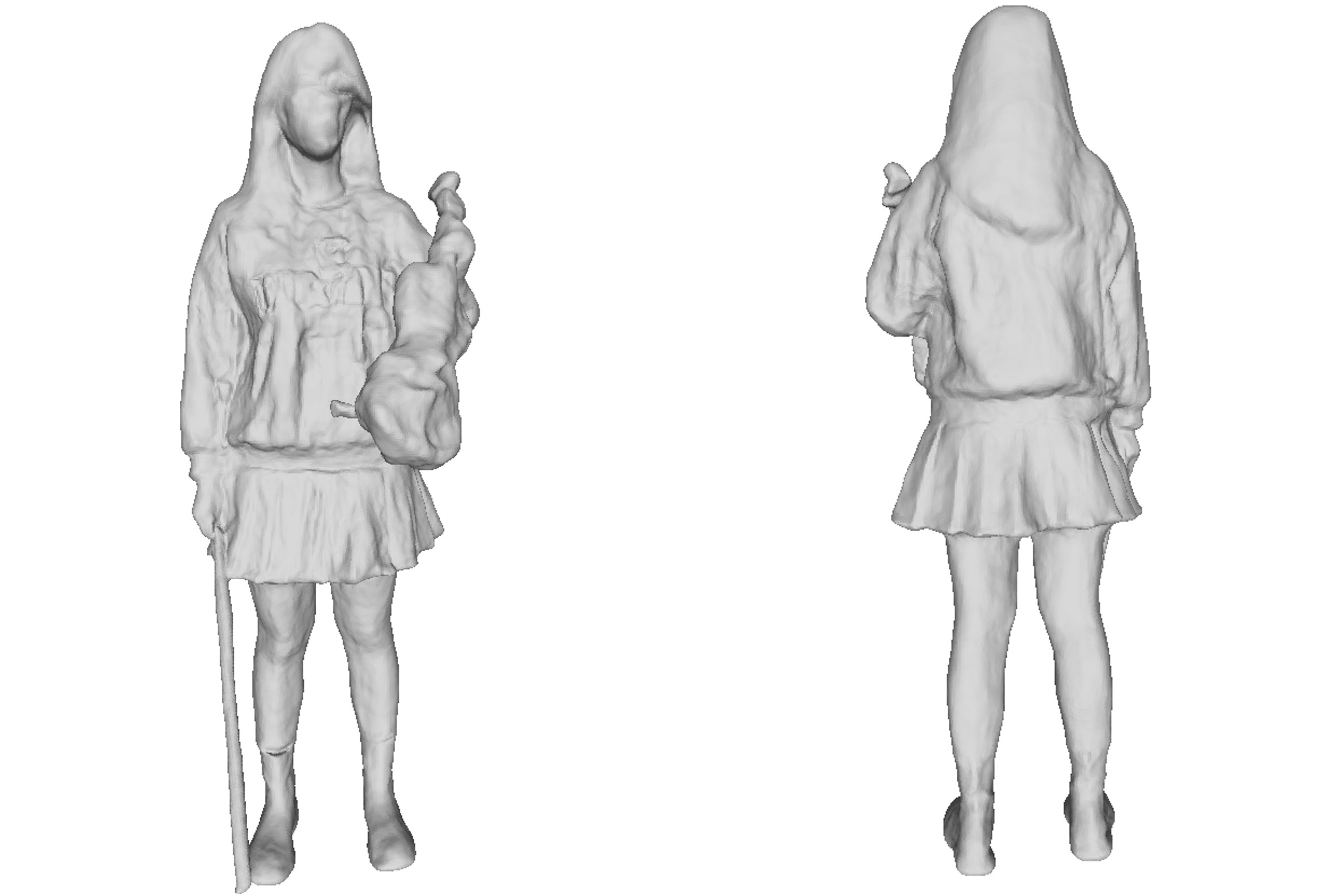}
    \includegraphics[width=0.162\linewidth]{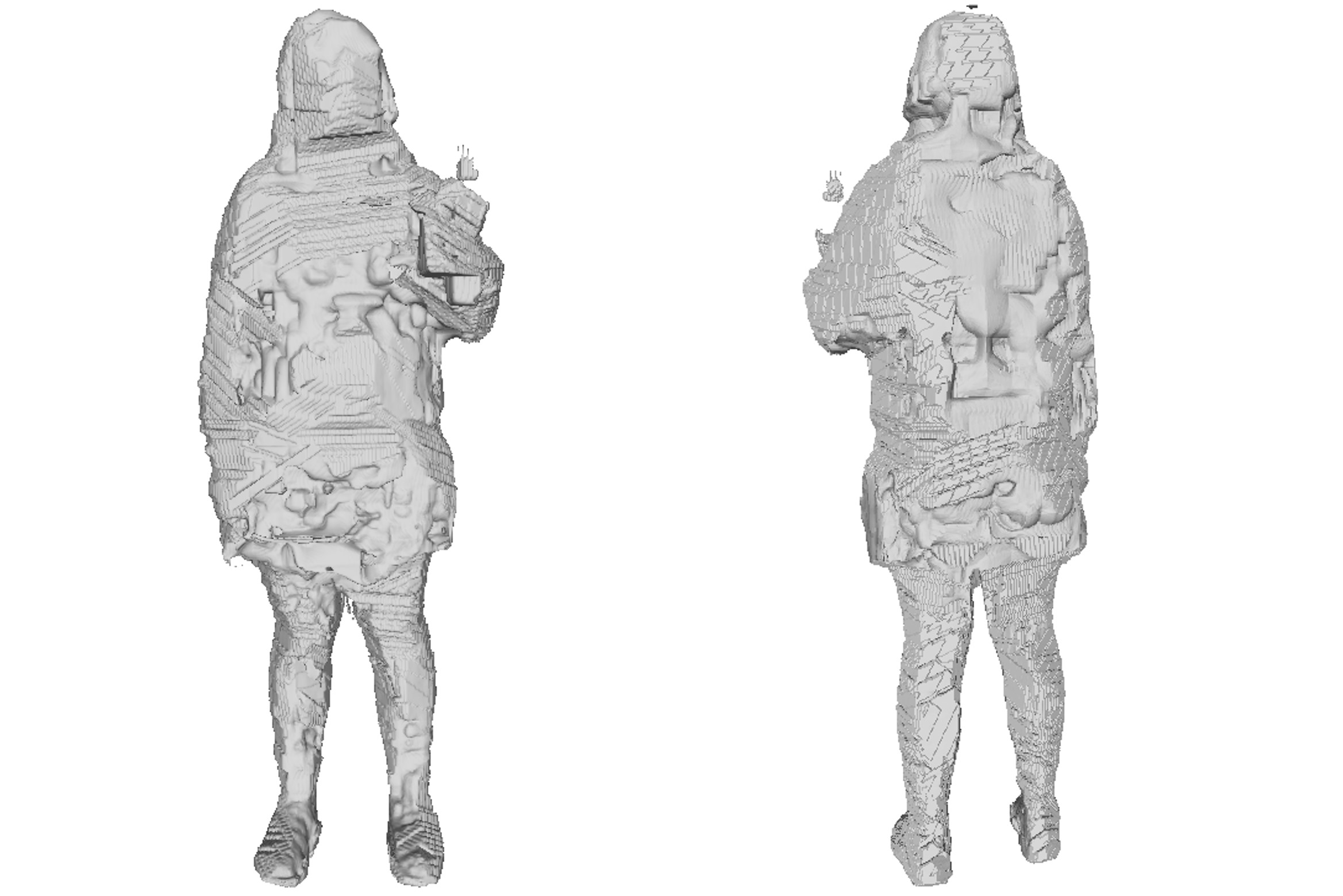}
    \includegraphics[width=0.162\linewidth]{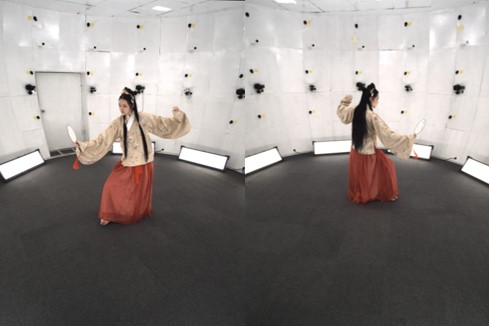}
    \includegraphics[width=0.162\linewidth]{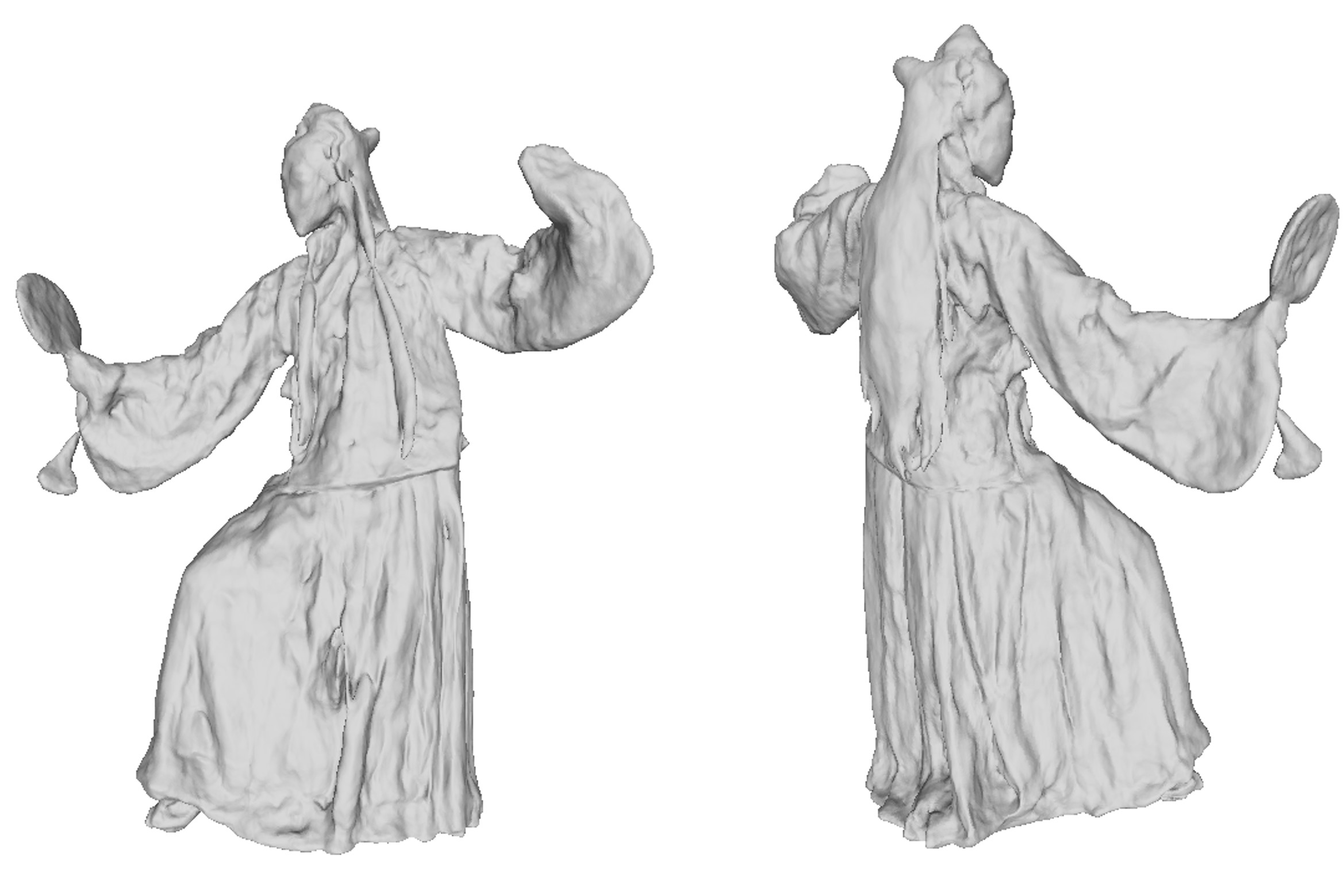}
    \includegraphics[width=0.162\linewidth]{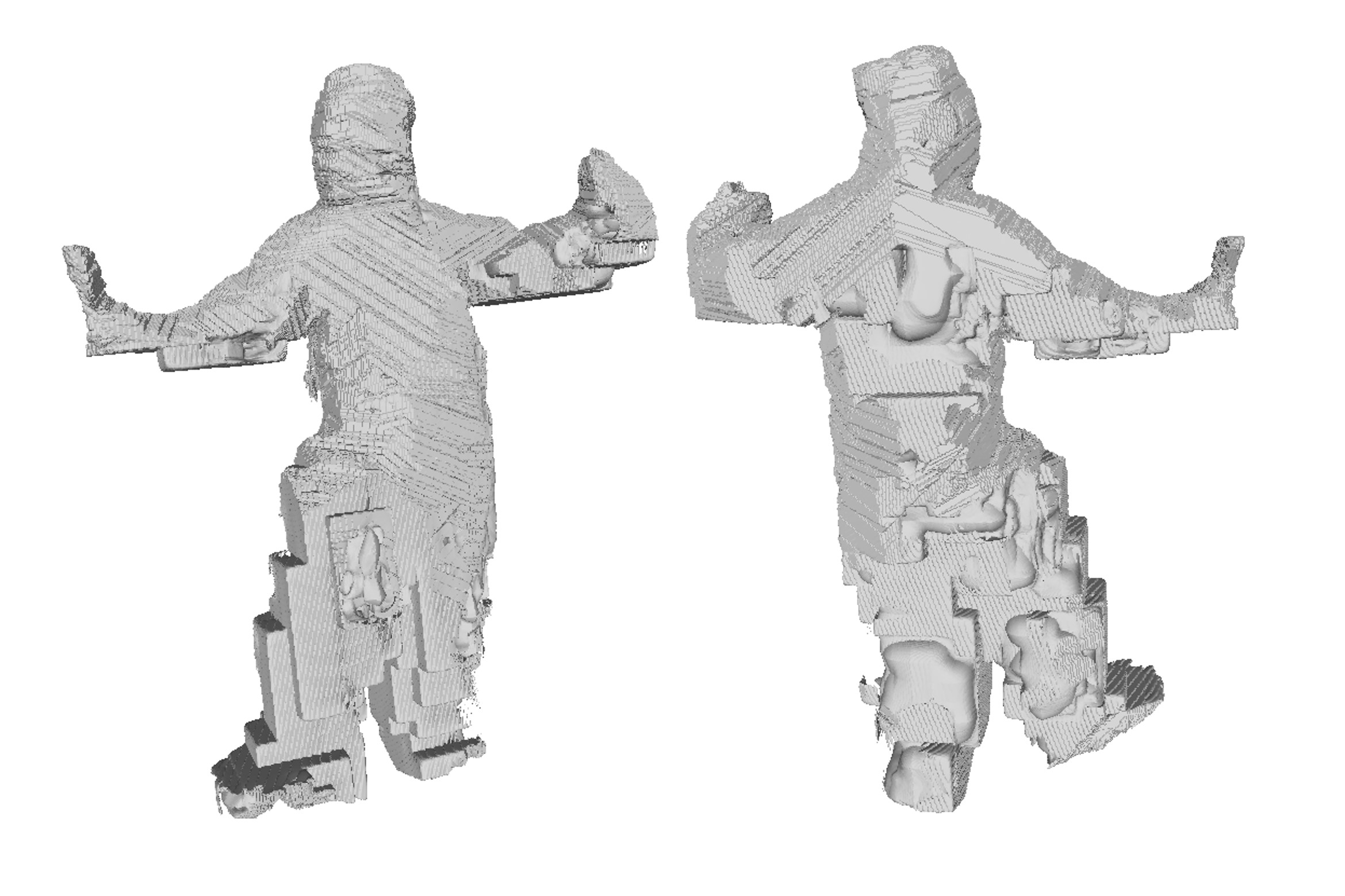}
    
    \captionsetup[subfigure]{labelformat=empty}
     \begin{subfigure}[b]{0.162\linewidth}
         \centering
         \includegraphics[width=\textwidth]{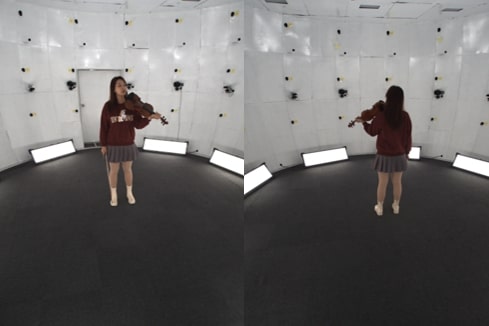}
        \caption{Input}
     \end{subfigure}
     \begin{subfigure}[b]{0.162\linewidth}
         \centering
         \includegraphics[width=\textwidth]{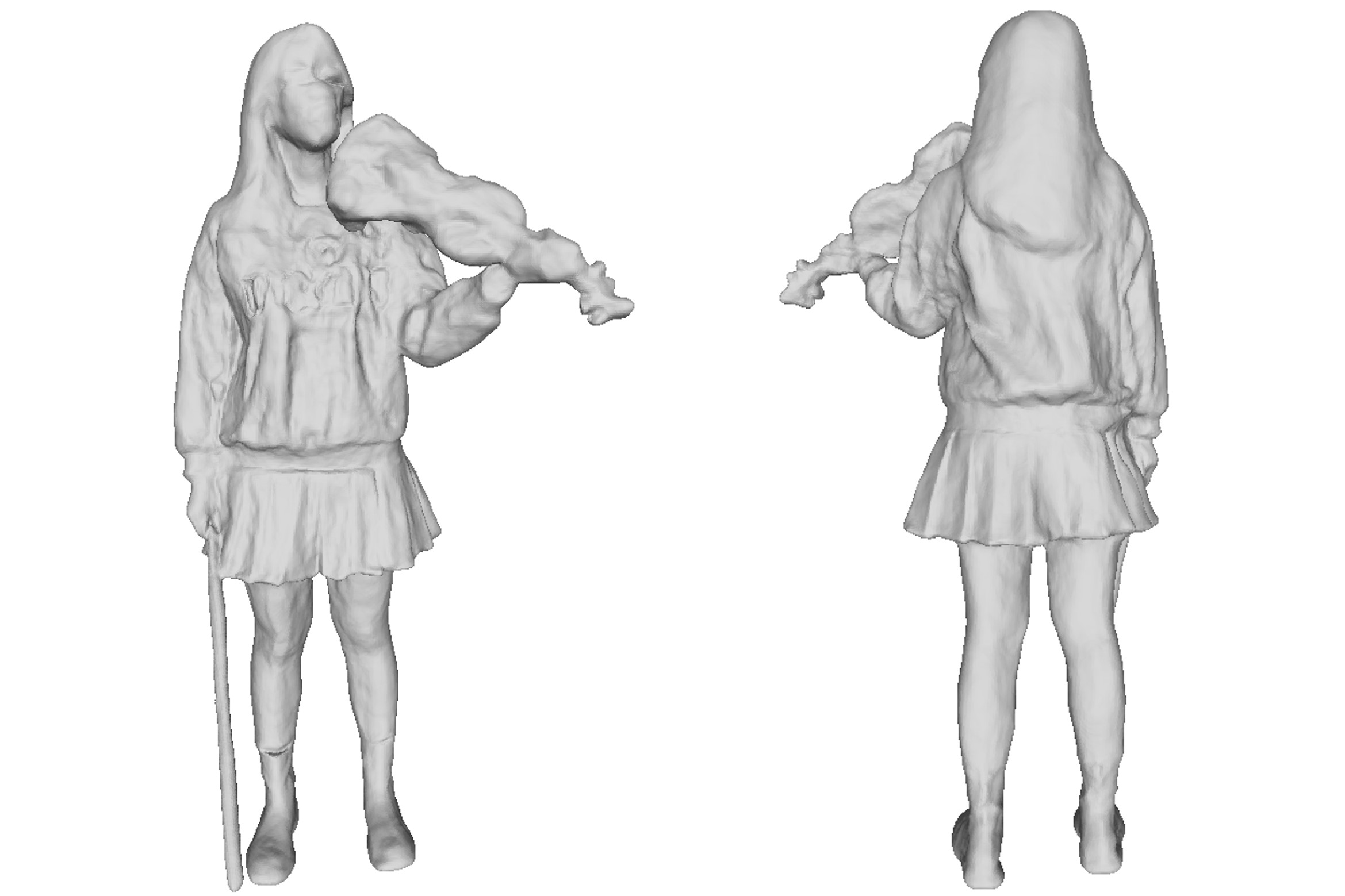}
        \caption{Ours}
     \end{subfigure}
     \begin{subfigure}[b]{0.162\linewidth}
         \centering
         \includegraphics[width=\textwidth]{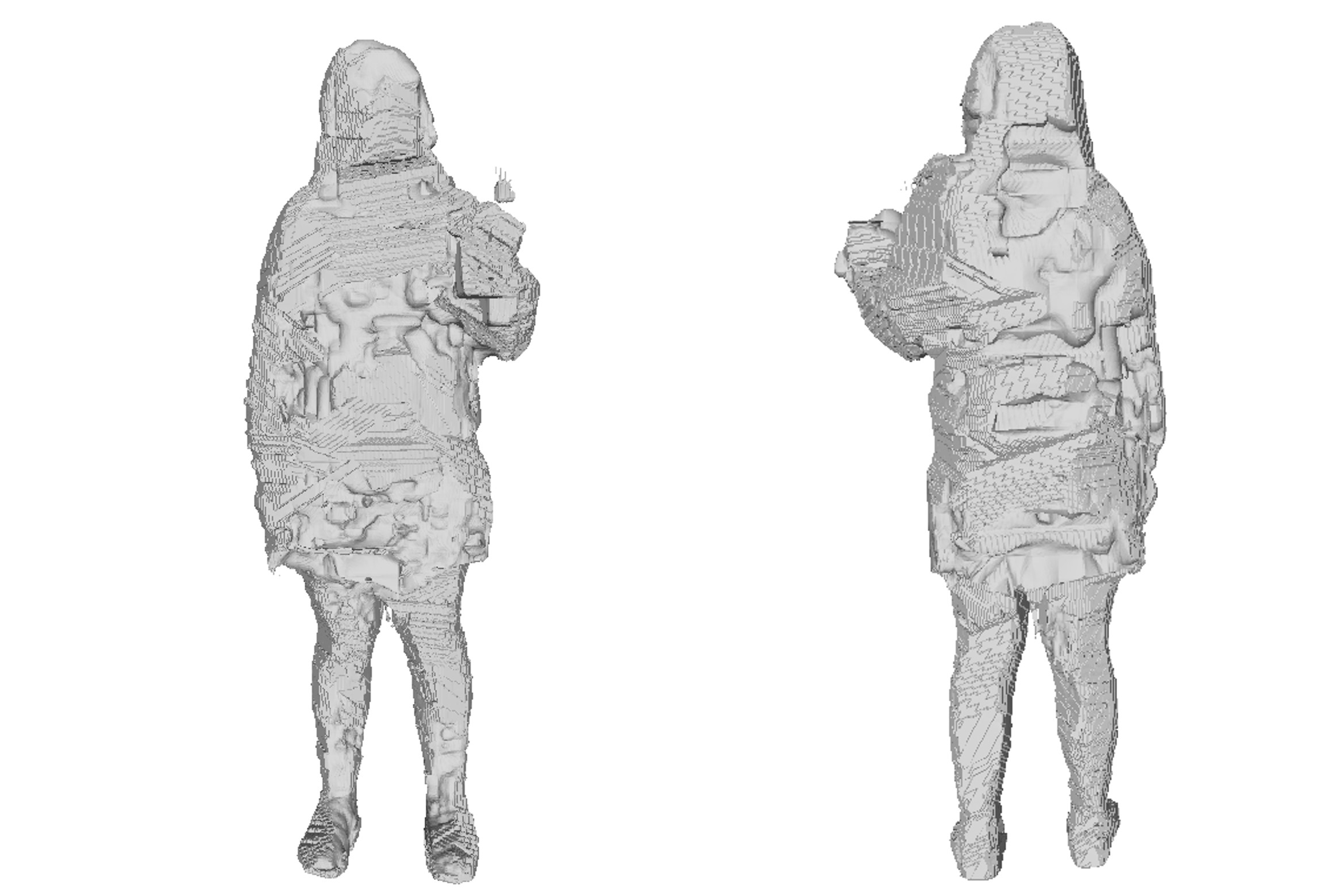}
        \caption{Neural Body}
     \end{subfigure}
     \begin{subfigure}[b]{0.162\linewidth}
         \centering
         \includegraphics[width=\textwidth]{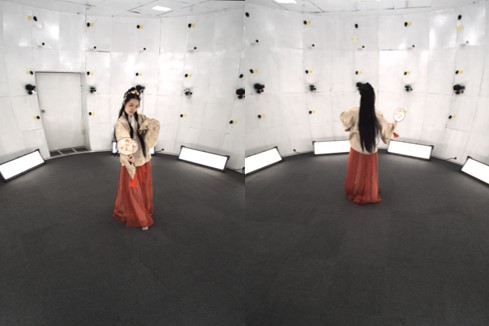}
        \caption{Input}
     \end{subfigure}
     \begin{subfigure}[b]{0.162\linewidth}
         \centering
         \includegraphics[width=\textwidth]{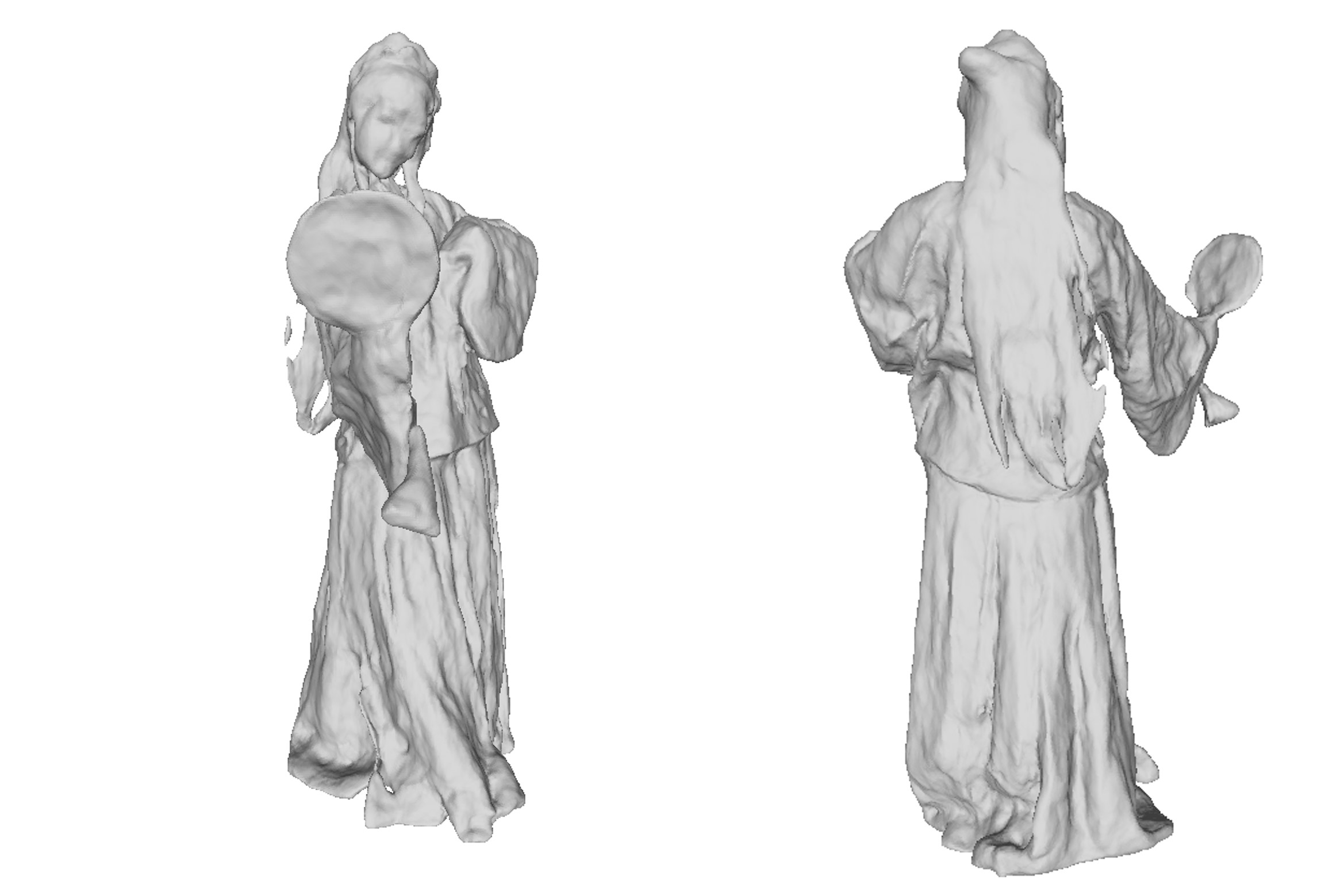}
        \caption{Ours}
     \end{subfigure}
     \begin{subfigure}[b]{0.162\linewidth}
         \centering
         \includegraphics[width=\textwidth]{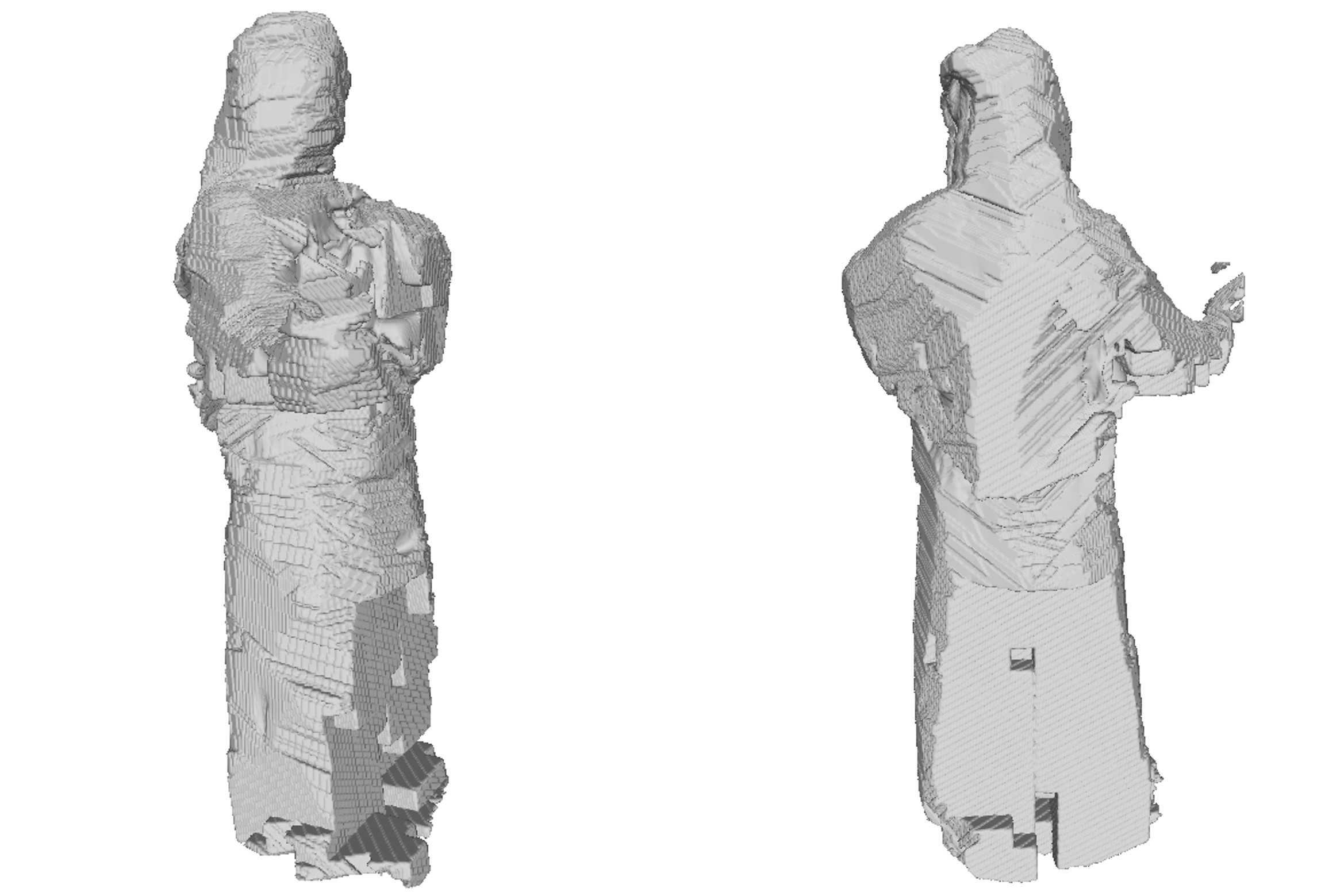}
        \caption{Neural Body}
     \end{subfigure}
    
    \caption{Dynamic surface reconstruction results on our collected dataset (the first 2 rows) and Genebody (the last 2 rows) datasets. For each scene, we show two views (horizontally aligned) and two frames (vertically aligned). }
    \label{fig:comparison}
\end{figure*}

\subsection{Mask-based Ray Selection} \label{selecion}
Previous implicit neural reconstruction methods in both monocular~\cite{d-nerf,nerfies,hypernerf} and multi-view~\cite{unisurf,neus,volsdf} sample a certain number of pixels uniformly over the whole input image to generate a batch for training. In this case, a large proportion of selected pixels may fall into the background, which usually dominates an image
but contributes little to the reconstruction result.
Therefore, we develop a novel masked-based ray selection algorithm for volume rendering to assign higher probability in the regions of interest, especially the time-varying foreground areas.


The key to the ray selection strategy is to design a probability map to guide the pixel sampling. One simple solution is dividing the probability map into two constant parts according to the segmentation mask at the current frame (e.g., Fig.~\ref{fig:mask}) with a higher value for the foreground region. We call this solution "naive ray selection" in Section~\ref{sec:abl}. However, 
the motion of foreground objects is generally not uniform. For instance, in Fig.~\ref{fig:img}, where a standing person is throwing up a ball, the movements of the ball and human hands are more significant than other parts of the foreground objects, and thus require more attention during optimization. Assuming the cameras are fixed over time, we propose a temporally global strategy to find the time-varying region over frames. Specifically, for each view $j$, we calculate a foreground frequency map $Q_j$ by summing up the segmentation masks over all $N$ frames before normalization:
\begin{equation}
    Q_j=\frac{1}{N}\sum_{i=1}^{N} S_{i,j},
\label{freq}
\end{equation}
where $ S_{i,j} \in \{0, 1\}^{H\times W}$ is the segmentation mask of the $i$-th frame and $j$-view. Then we derive a dynamic motion map $D_j$ 
from:
\begin{equation}
    D_j=\mu_{max} - (\mu_{max}-\mu_{min})Q_j.
\label{dynamicmotionmap}
\end{equation}
Since the frequency map $Q_j$ is normalized to the interval $[0,1]$, the dynamic motion map $D_j$ ranges from $\mu_{min}$ to $\mu_{max}$ while higher values represent more significant motion (see an example in Fig.~\ref{fig:motion}).

Similarly, we also differentiate the background area. In our method, the rays in the background area are supervised by a binary cross-entropy loss to carve the empty space along the rays. 
Compared to the background regions which are far from the foreground objects, we argue that the closer areas are of more interest. 
To this end, we generate a buffer region in the background by morphologically dilating the object mask with a radius of $r_{dilate}$ before subtracting the original mask by the dilated one. Since this buffer region deserves more attention than the rest of the background, we assign it with a higher probability for ray selection. 

Finally, the probability map of the $i$-th frame and the $j$-th view $P_{i,j}$ for sampling rays is defined as:
\begin{equation}
    P_{i,j}(\mathbf{p})=\left\{
    \begin{split}
        &D_j(\mathbf{p}) &\text{if} \; \mathbf{p} \in \mathcal{F}_{i,j}\\
        &\mu_{buffer} &\text{if} \; \mathbf{p} \in \mathcal{B}_{i,j}\\
        &\mu_{rest} &\text{if} \; \mathbf{p} \in \mathcal{R}_{i,j}
    \end{split}
    \right. ,
\label{probabilitymap}
\end{equation}
where $\mathbf{p}$ denotes a pixel, while $\mathcal{F}_{i,j}$, $\mathcal{B}_{i,j}$ and $\mathcal{R}_{i,j}$ are the sets of pixels in the foreground, buffer and the remaining background regions, respectively.
Note that the values in the probability map serve as the pixel sampling weights, and will be further normalized over the whole map to compute the final sampling probability. 
Therefore, only the relative values rather than the absolute ones between $\mu_{max}$, $\mu_{min}$, $\mu_{buffer}$ and $\mu_{rest}$ affect the ray selection.
As shown in Figure~\ref{fig:prob_map}, the time-varying regions, including the ball and hands, are of higher interest inside the foreground region, while the buffer area in the background is also given more attention than the rest.

\subsection{Loss Function} \label{optimization}
Each batch of training data consists of $N_{ray}$ rays sampled from one single image $I_{i,j}$ of the $i$-th frame and the $j$-th view. For simplicity, we drop the indices $i$ and $j$ in this section. 
We first minimize the difference between the rendered colors in the foreground and the ground truth ones, denoted by $\mathcal{L}_{rgb}$.
Then we compute a mask loss to supervise the geometry field in canonical space:
\begin{equation}
\begin{aligned}
    \mathcal{L}_{mask} = \frac{1}{N_{ray}}\sum_{\mathbf{r} \in \mathcal{R}} \text{BCE}(\hat{S}(\mathbf{r}), S(\mathbf{r})),
\end{aligned}
\end{equation}
where $\hat{S}$ is the volume rendered mask and BCE is the binary cross-entropy loss. 
Similar to the elastic regularization in ~\cite{nerfies}, we regularize the local deformations by encouraging a rigid motion using $\mathcal{L}_{rigid}$.
Lastly, we adopt the Eikonal loss~\cite{igr} $\mathcal{L}_{eikonal}$ to regularize the SDF gradients (i.e., surface normals $\mathbf{n}$) of both $\mathcal{X}$ and an additional set of sample points $\mathcal{P}$ distributed uniformly in the bounding volume. 
Finally, the total loss function is formed as:
\begin{equation}
    \mathcal{L}_{total} = \mathcal{L}_{rgb} + \lambda_1 \mathcal{L}_{mask} + 
     \lambda_2 \mathcal{L}_{rigid} + \lambda_3 \mathcal{L}_{eikonal},
\end{equation}
where $\lambda_1, \lambda_2, \lambda_3$ are the weights for respective term.

\section{Experiments}
\subsection{Experimental Settings} \label{setup}
\noindent
\textbf{Implementation Details.}
The networks of $T, H, F, R$ are MLPs containing seven, seven, nine, and five fully-connected layers, respectively.
We empirically set the number of hyper-coordinates $m$ as 2. The deformation code $\varphi_i$ and appearance codes $\psi_i, \chi_j$ all have 8 dimensions. And we set the frequencies of positional encoding~\cite{nerf} as 6, 4, and 1 for spatial coordinates in observation and canonical space, view direction, and hyper-coordinates, respectively. In each training iteration, a batch contains $N_{ray}=512$ rays, along which $N_s=128$ spatial points are sampled. The loss weights are empirically set as $\lambda_1=1.0, \lambda_2=0.05, \lambda_3=0.1$. Adam~\cite{adam} is adopted for training, with the learning rate of $2.5\times 10^{-4}$ for the deformation network $T$ and $5\times 10^{-4}$ for the rest of the parameters.
For generating probability maps for ray selection, we set $r_{dilate}$, $\mu_{max}$, $\mu_{min}$, $\mu_{buffer}$ and $\mu_{rest}$ as 80, 1, 0.3, 0.1, 0.001, respectively.
After training, surface meshes are extracted via Marching Cubes~\cite{lorensen1987marching} at a resolution of $512^3$ voxels.
We conduct the training on one single NVIDIA GTX1080Ti GPU. The training on a video of 100 frames takes about 300k iterations (around 70 hours) to converge. 


\noindent
\textbf{Datasets.} 
To evaluate the performance of our method, we train our model on both Genebody~\cite{cheng2022genebody} and our datasets. They consist of human performers of various ages, intricate gestures, clothing types, and accessories. In particular, Genebody~\cite{cheng2022genebody} is captured by 48 evenly spaced cameras with a resolution of $2448\times 2048$, while we only take 16 out of 48 cameras for our experiments. 
In addition, we collect a volumetric video dataset 
by 16 uniformly distributed cameras at resolution of $5120\times 3840$ and 25 fps, while only the down-sampled images ($1280\times 960$) are used for training. 
Both datasets consist of human performers of various ages, intricate gestures, clothing types, and accessories.All sequences have a length between 100 and 150 frames.

\begin{figure}[t]
    \centering
    \captionsetup[subfigure]{labelformat=empty}

     \begin{subfigure}[b]{0.325\linewidth}
         \centering
         \includegraphics[trim={10cm 4cm 3cm 3cm},clip,width=0.485\textwidth, height=0.958\textwidth] {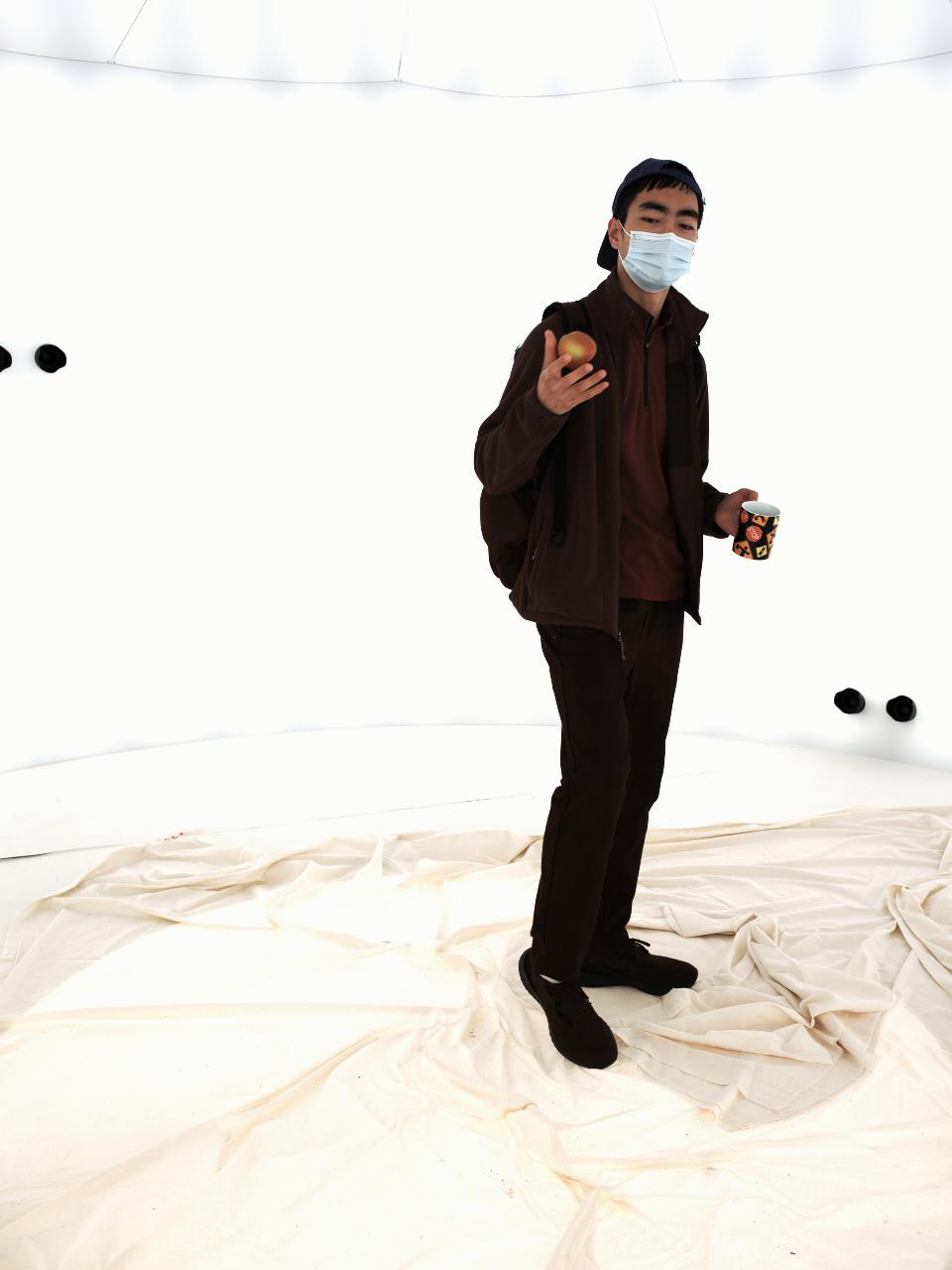}
         \includegraphics[trim={36cm 22cm 34cm 18cm},clip,width=0.485\textwidth] {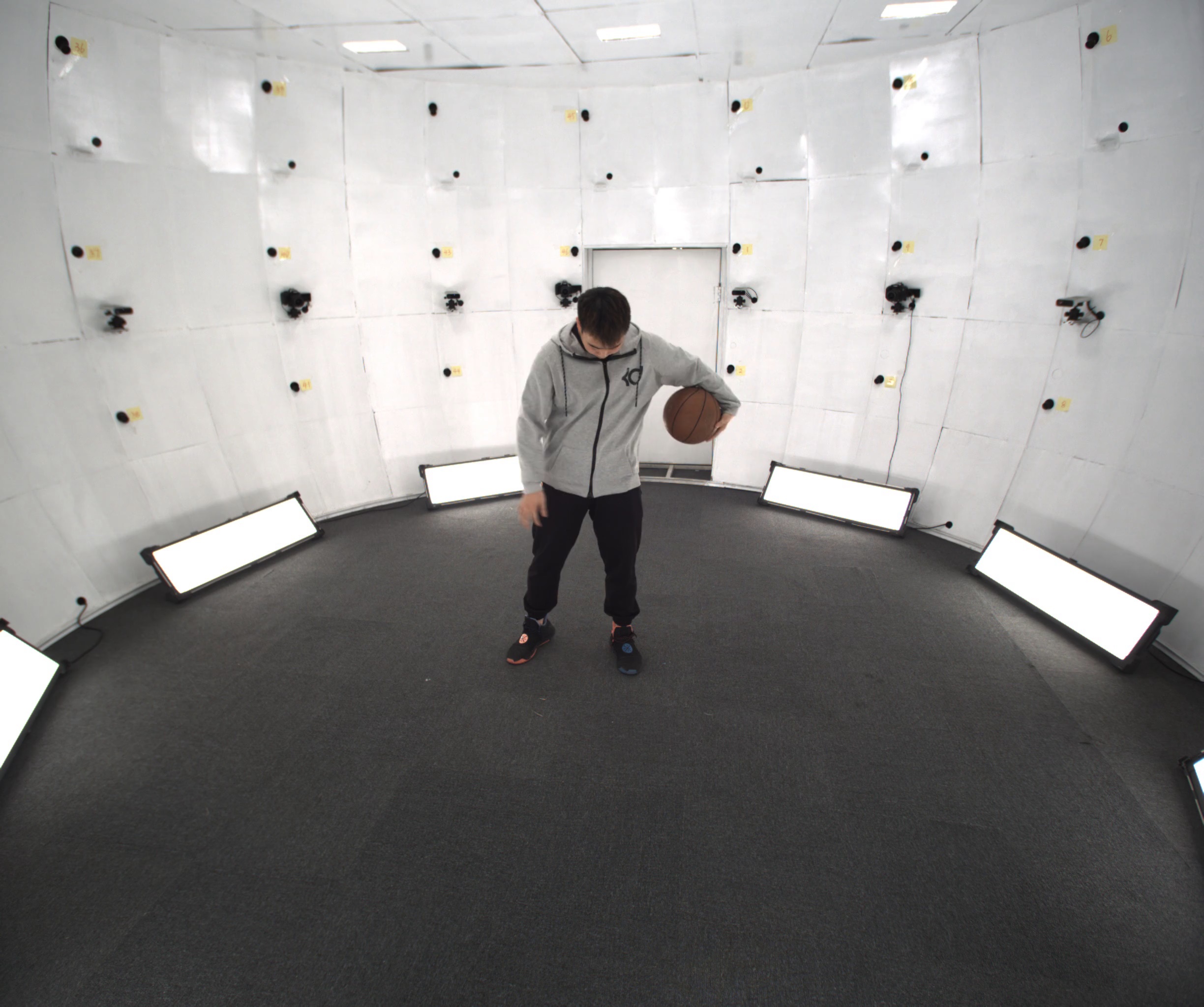}
        \caption{Reference}
     \end{subfigure}
    \hfill
     \begin{subfigure}[b]{0.325\linewidth}
         \centering
         \includegraphics[trim={10cm 5cm 3cm 4cm},clip,width=0.485\textwidth, height=0.958\textwidth] {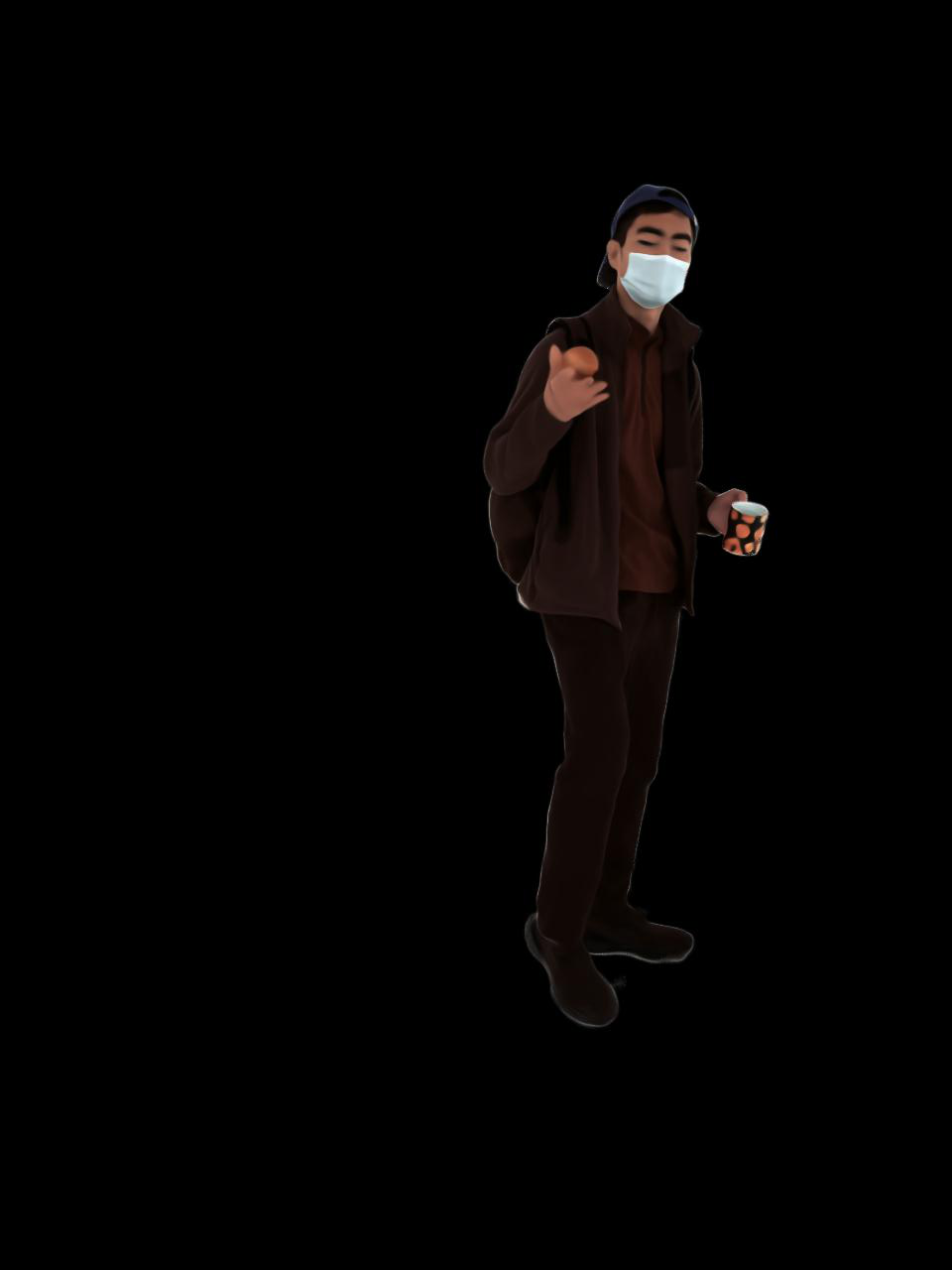}
         \includegraphics[trim={9cm 5.5cm 8.5cm 4.5cm},clip,width=0.485\textwidth] {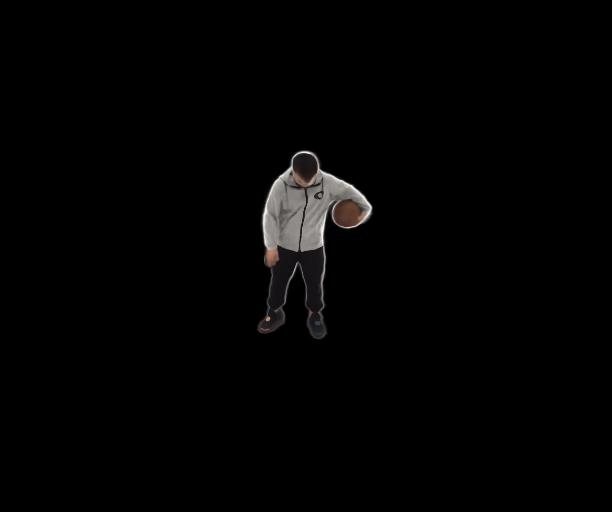}
        \caption{Ours}
     \end{subfigure}
    \hfill
     \begin{subfigure}[b]{0.325\linewidth}
         \centering
         \includegraphics[trim={13cm 5cm 3.5cm 4cm},clip,width=0.485\textwidth, height=0.958\textwidth] {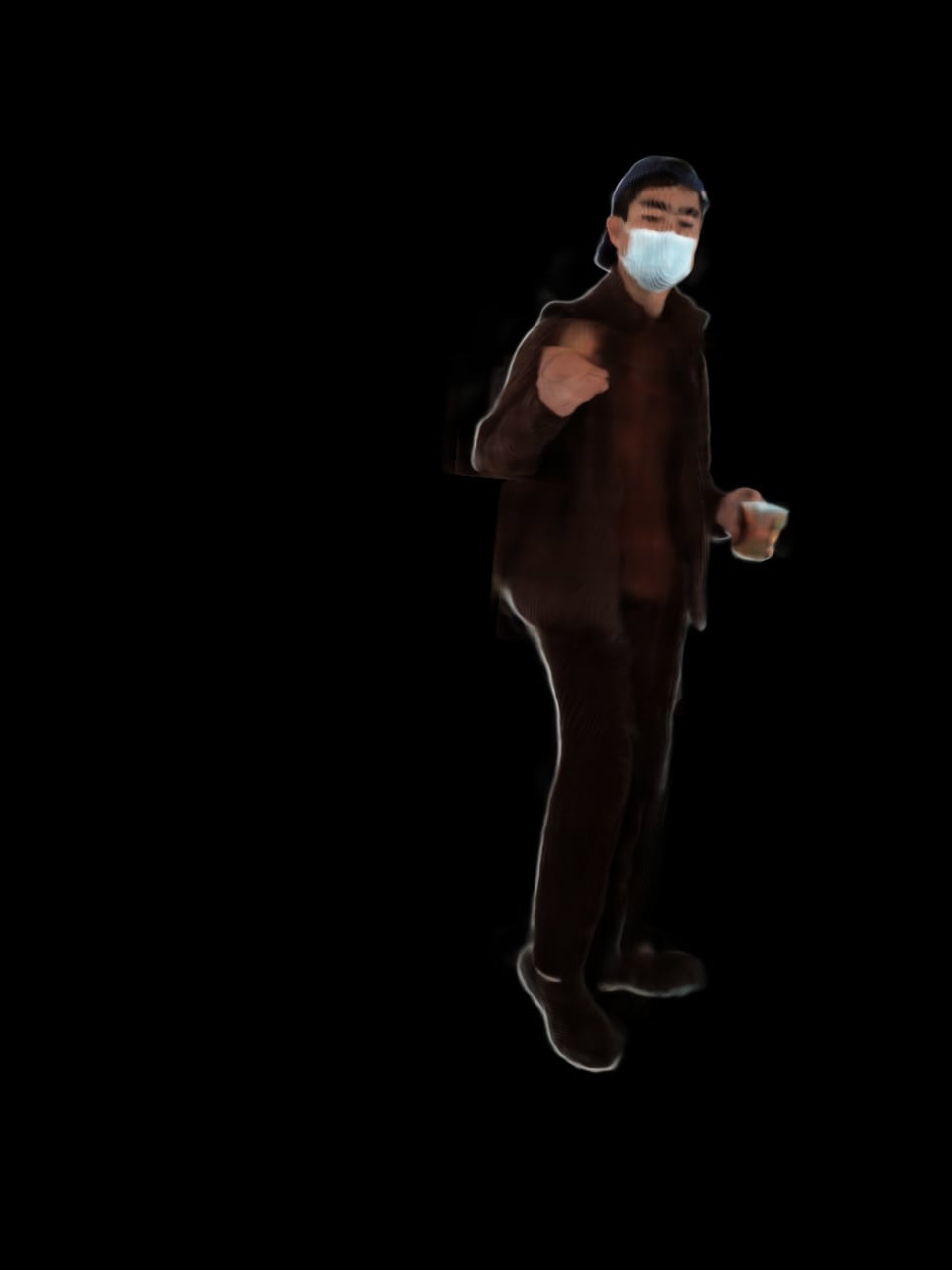}
         \includegraphics[trim={36cm 22cm 34cm 18cm},clip,width=0.485\textwidth] {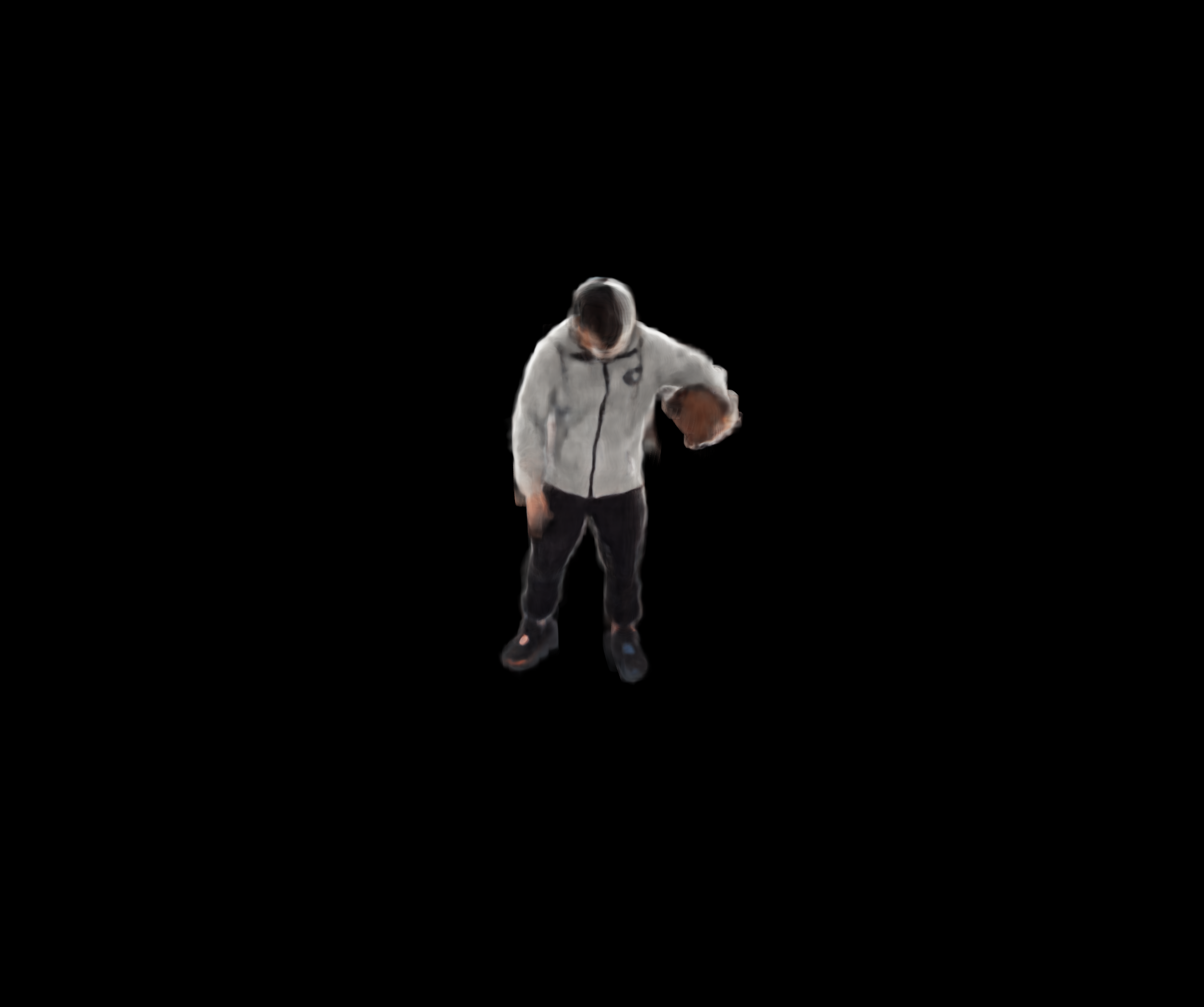}
        \caption{Neural Body}
     \end{subfigure}
    
    \caption{Qualitative comparison of novel view synthesis.}
    \label{fig:viewsyn}
\end{figure}

\subsection{Comparison}\label{res}
Since most of state-of-the-art multi-view dynamic reconstruction approaches are template-based, we have evaluated Neural Body~\cite{peng2021neural}, A-Nerf~\cite{anerf} and AniSDF~\cite{peng2022animatable} on both mentioned datasets as they are also human-driven dynamic scenes, following the official codes and instructions. However, we found that the performance of the last two methods on the above datasets is unsatisfying, so we only compare our method with Neural Body in this paper. 

Figure~\ref{fig:comparison} demonstrates qualitative results for dynamic surface reconstruction. 
While Neural Body~\cite{peng2021neural} fails to recover objects beyond human bodies and loose dresses,
our method is robust to complicated motions with topological changes (e.g., taking off a jacket), and reconstruct high-fidelity surface meshes even for challenging thin objects, including the violin bow and the round handheld fan. In addition, our method requires no prior knowledge on the scene objects like parametric templates and thus is able to reconstruct general dynamic scenes, including loose clothing and various props.

Figure~\ref{fig:viewsyn} shows a qualitative comparison on rendering novel views (excluded from the training) on different datasets. The rendering results from Neural Body suffers from the ghost effect and missing parts of objects (e.g., the basketball). In contrast, our method synthesizes photo-realistic images from novel views with more appearance details. 
To measure the rendering quality on testing views not used for training, we choose three metrics: peak signal-to-noise ratio (PSNR), structural similarity index (SSIM) and LPIPS~\cite{lpips}. Following previous works~\cite{peng2021neural,peng2022animatable}, we set the values of the background pixels as zero. Instead of evaluating the whole images, which leads to meaningless high scores, we crop images using minimal bounding boxes of the corresponding foregrounds and only calculate the metrics on the cropped regions. As illustrated in Table~\ref{table:viewsyn}, our method outperforms Neural Body with a large margin on novel view synthesis on both our collected dataset (Volumetric Videos) and Genebody dataset.

\begin{table}[t]
\begin{center}
\resizebox{1\linewidth}{!}{
\begin{tabular}{c|ccc|ccc}
  &  \multicolumn{3}{c|}{Volumetric Videos} & \multicolumn{3}{c}{Genebody} \\
  & PSNR$\uparrow$ & SSIM$\uparrow$ & LPIPS$\downarrow$  & PSNR$\uparrow$ & SSIM$\uparrow$ & LPIPS$\downarrow$ \\
\hline 
Neural Body  & 26.84          & 0.894          & 0.246          & 19.87         & 0.723           & 0.234 \\
\hline 
No HC; no RS & 25.83          & 0.872          & 0.280          & 20.25         & 0.714           & 0.252 \\
No RS        & 27.44          & 0.902          & 0.178          & 24.81         & 0.803           & 0.189 \\
Naive RS     & 27.26          & 0.909          & 0.175          & 25.15         & 0.820           & 0.171 \\
Ours         & \textbf{28.33} & \textbf{0.923} & \textbf{0.143} & \textbf{26.66} & \textbf{0.841} & \textbf{0.145}
\end{tabular}
}
\end{center}
\caption{Quantitative results of novel view synthesis.}
\label{table:viewsyn}
\end{table}

\begin{figure}[t]
     \centering
     \begin{subfigure}[b]{0.35\columnwidth}
         \centering
         \includegraphics[width=\textwidth]{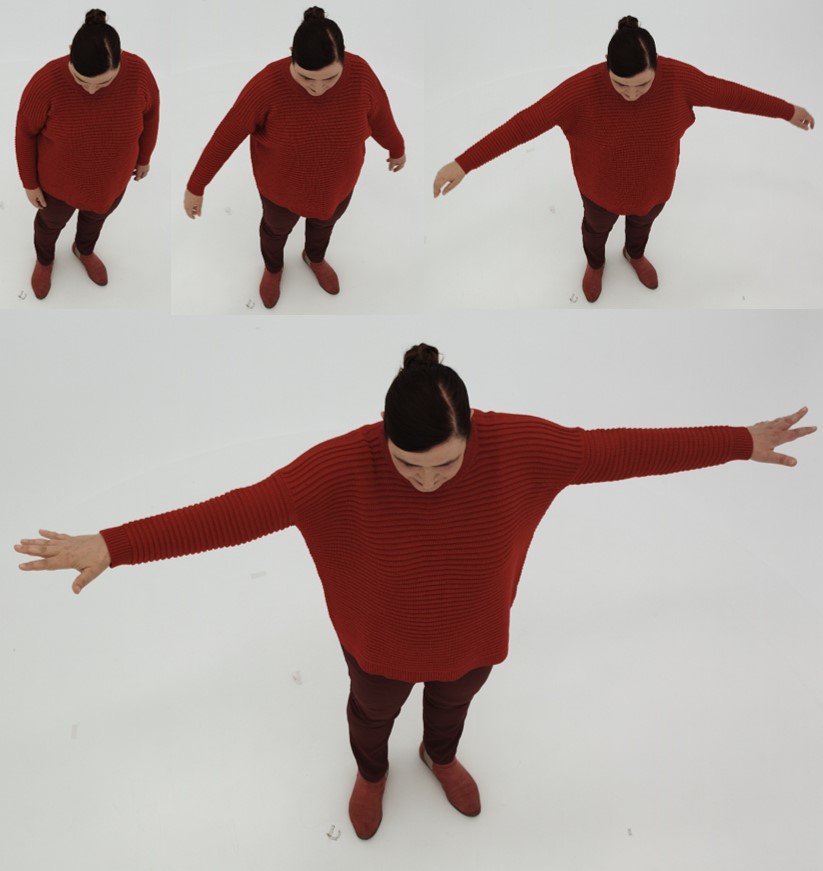}
         \caption{Input examples}
         \label{fig:input}
     \end{subfigure}
    \hspace*{\fill}
     \begin{subfigure}[b]{0.245\columnwidth}
         \centering
        \captionsetup{justification=centering}
         \includegraphics[trim={13mm 10mm 13mm 0mm},clip, width=\textwidth] {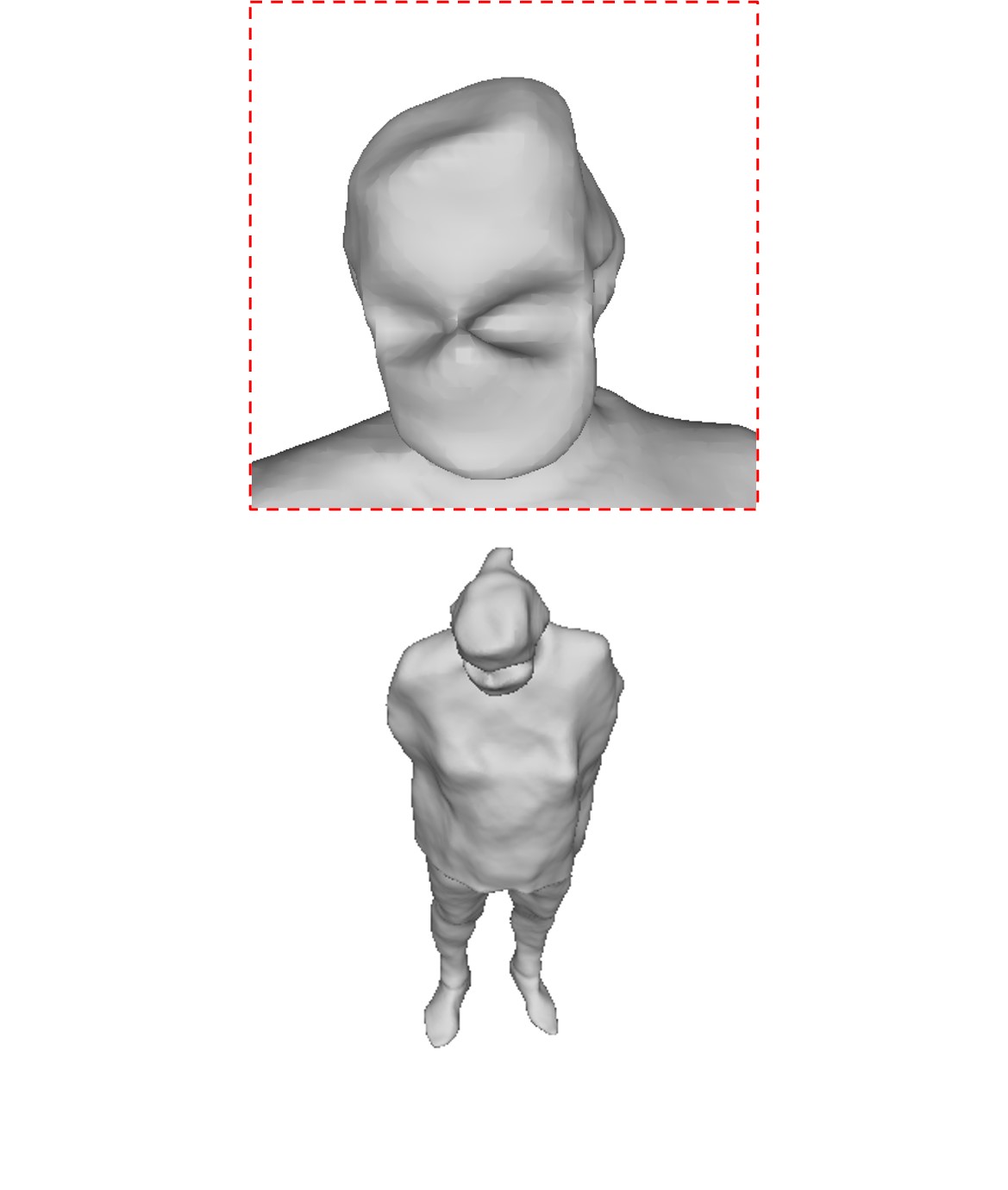}
        \caption{No HC; no RS}
         \label{fig:noh}
     \end{subfigure}
    \hspace*{\fill}
     \begin{subfigure}[b]{0.35\columnwidth}
         \centering
        \captionsetup{justification=centering}
         \includegraphics[trim={0mm 10mm 0mm 0mm},clip, width=\textwidth]{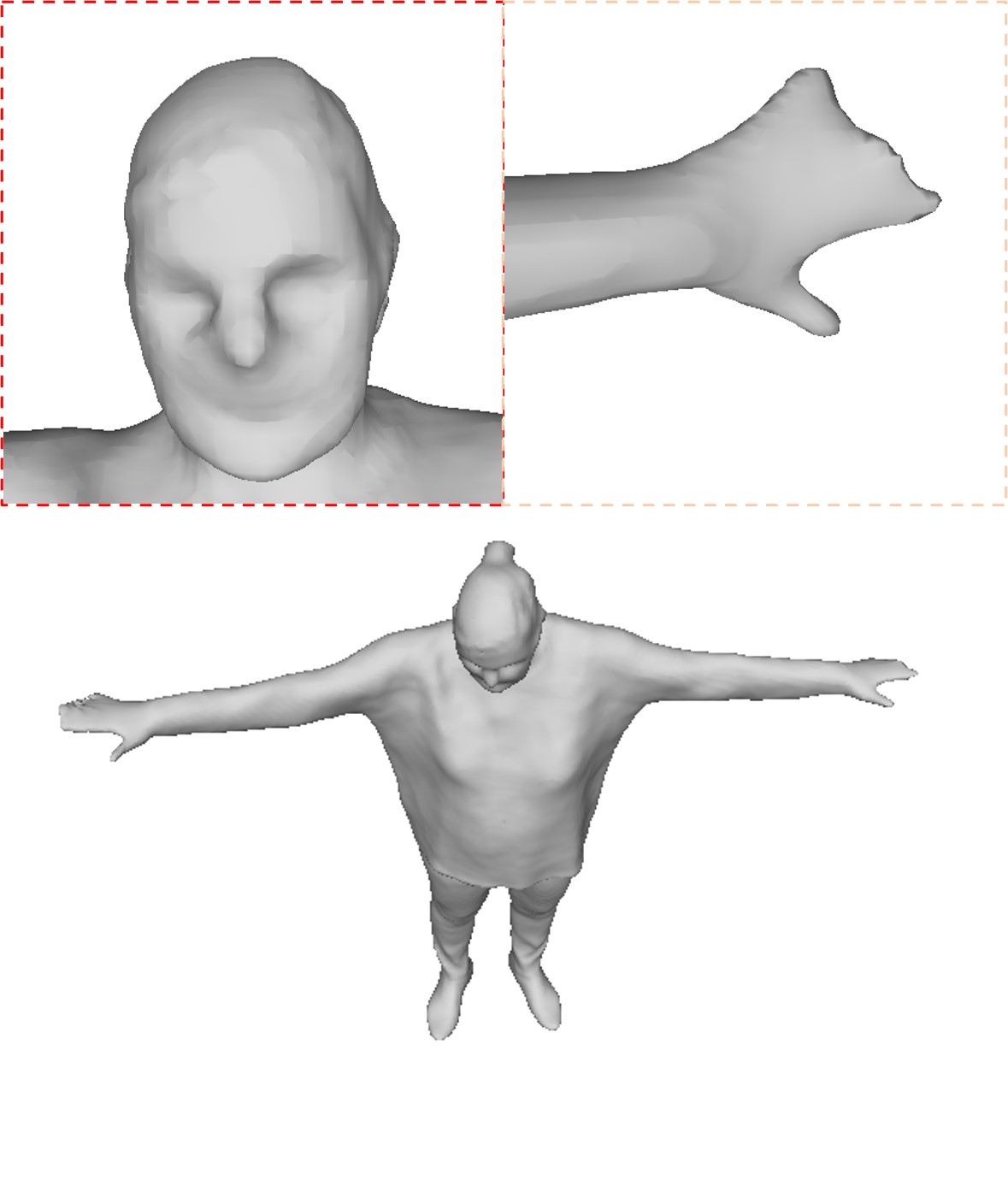}
         \caption{No RS}
         \label{fig:wors}
     \end{subfigure}
    \hspace*{\fill}
    
    \vspace{2mm}
    
    \hspace*{\fill}
     \begin{subfigure}[b]{0.35\columnwidth}
         \centering
        \captionsetup{justification=centering}
         \includegraphics[trim={0mm 10mm 0mm 0mm},clip, width=\textwidth]{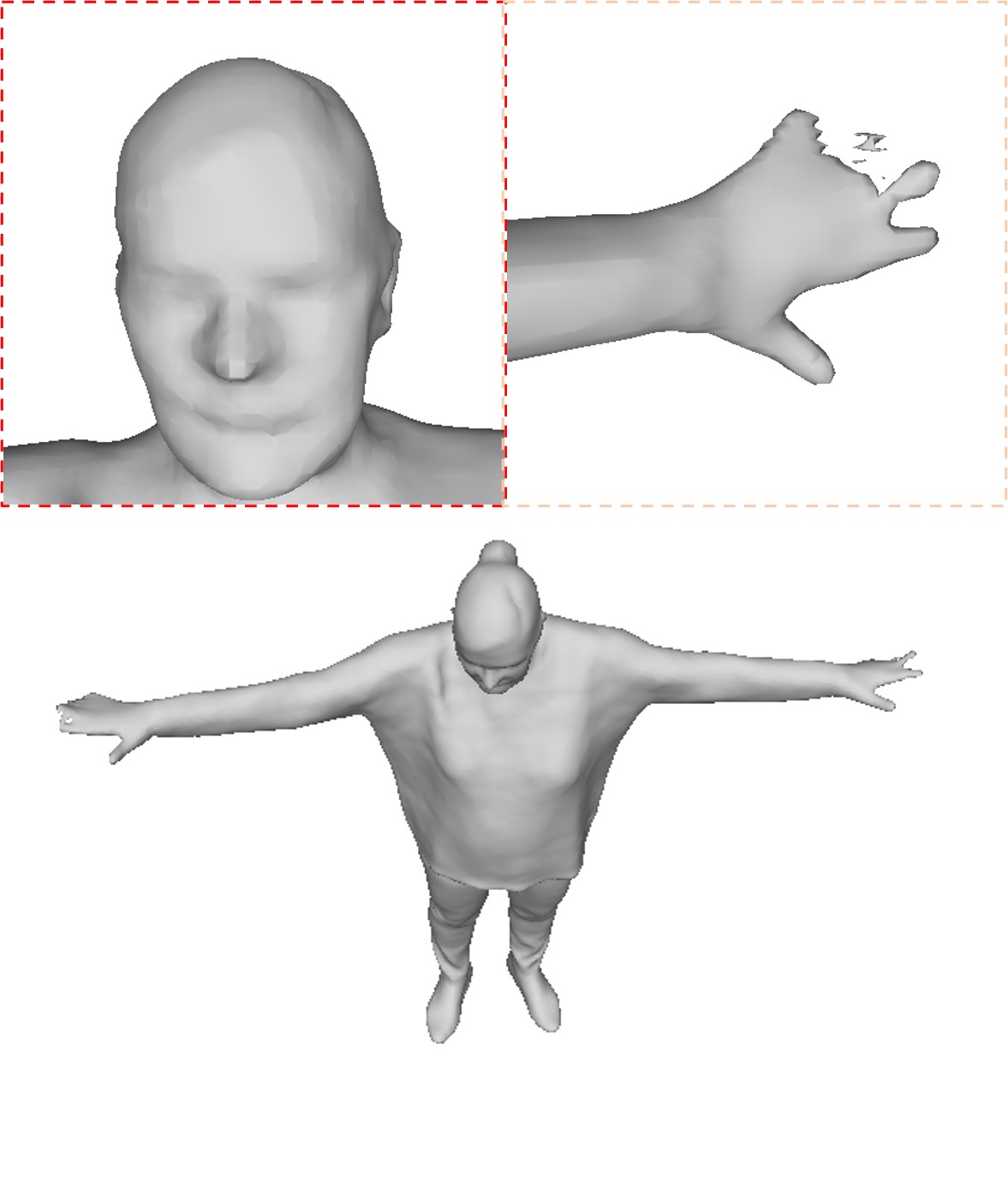}
         \caption{Naive RS}
         \label{fig:st}
     \end{subfigure}
    \hspace*{\fill}
     \begin{subfigure}[b]{0.35\columnwidth}
         \centering
        \captionsetup{justification=centering}
         \includegraphics[trim={0mm 10mm 0mm 0mm},clip, width=\textwidth]{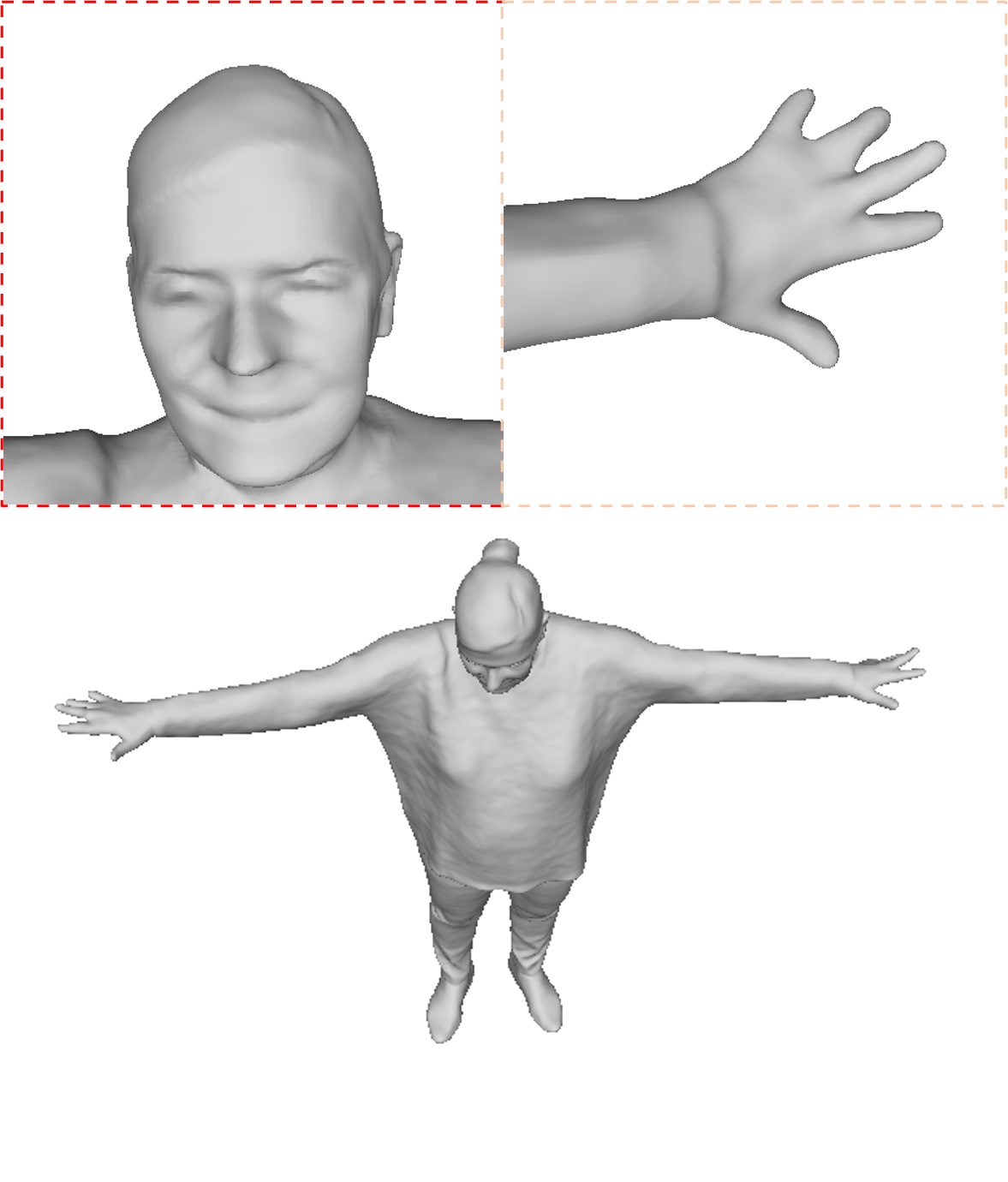}
         \caption{Ours}
         \label{fig:dy}
     \end{subfigure}
    \hspace*{\fill}
\caption{Ablation of hyper-coordinates (HC) and ray selection (RS).}
\label{fig:abl}
\end{figure}

\subsection{Ablation Study}
\label{sec:abl}
We present an ablation study to report the effect of the important components of our method on the final reconstruction performance. Figure~\ref{fig:abl} demonstrates the reconstruction results with different choices between hyper-coordinates (HC) and ray selection (RS) strategies. Fig.~\ref{fig:noh} shows that ablating the hyper-coordinates network (i.e., representing the canonical space by only 3D instead of adding more dimensions) fails to fully recover both arms due to their topological changes resulting from situations where arms and the body touch or release each other.

The evaluation of the ray selection shows, that uniformly sampling rays over the whole image causes an imbalanced training: the model is trained well for most parts of the body that remain approximately static over time, but the geometric details of the moving hands and head are missing (see Fig.~\ref{fig:wors}). As shown in Fig.~\ref{fig:st}, the naive ray selection strategy improves the result by increasing the weights for the foreground, while it has no attention on the time-varying regions. In comparison, our temporally global ray selection strategy focuses on optimizing the dynamic regions and differentiates the background area, thus faithfully recovering the challenging moving objects, such as the fingers and the face (see figure~\ref{fig:dy}). In addition to surface reconstruction, our ray selection strategy also achieve the best rendering quality, as illustrated in Table~\ref{table:viewsyn}.


\section{Conclusions}
We have introduced DySurf, a template-free method for reconstruction of general dynamic scenes from multi-view videos using neural implicit surface representation. We employ a deformation field to warp observed frames into a static hyper-canonical space, which is jointly optimized with an SDF network and a radiance network through volume rendering. Our novel ray selection strategy allows to specifically train the time-varying regions of interest. This significantly improves the reconstruction quality. Extensive experiments show that our method outperforms a state-of-the-art template-based method on 
different datasets, achieving high-quality geometry reconstruction as well as photorealistic novel view synthesis.

\section{Acknowledgement}
This work has partly been funded by the H2020 European project Invictus under grant agreement no. 952147 as well as by the Investitionsbank Berlin with financial support by European Regional Development Fund (EFRE) and the government of Berlin in the ProFIT research project KIVI.


\bibliographystyle{IEEEbib}
\bibliography{refs}


\end{document}